\PassOptionsToPackage{table}{xcolor}
\documentclass{article} 
\usepackage{iclr2027_conference,times}

\usepackage[utf8]{inputenc} 
\usepackage[T1]{fontenc}    
\usepackage{hyperref}       
\usepackage{url}            
\usepackage{booktabs}       
\usepackage{amsfonts}       
\usepackage{nicefrac}       
\usepackage{microtype}      
\usepackage[table]{xcolor}  
\usepackage{multirow}       
\usepackage{array}          
\usepackage{tabularx}       

\usepackage{graphicx}       
\usepackage{amsmath}        
\usepackage{amsthm}         
\usepackage{bm}             
\usepackage{algorithm}      
\usepackage{algorithmic}    
\usepackage{xspace}
\usepackage{tcolorbox}      

\newcolumntype{C}[1]{>{\centering\arraybackslash}m{#1}}
\newcolumntype{L}[1]{>{\raggedright\arraybackslash}m{#1}}
\newcolumntype{Y}{>{\centering\arraybackslash}X}

\newcommand{\mname}{CurvDelta\xspace}
\newcommand{\mnamedense}{CurvDelta\xspace}
\newcommand{\mnamelora}{CurvLoRA\xspace}

\newtheorem{theorem}{Theorem}
\newtheorem{proposition}[theorem]{Proposition}

\theoremstyle{definition}
\newtheorem{assumption}{Assumption}
\theoremstyle{remark}

\theoremstyle{plain}

\definecolor{basebg}{HTML}{EEEEEE}     
\definecolor{methodbg}{HTML}{E3F2FD}    
\definecolor{avgbg}{HTML}{F2F0EB}      

\definecolor{heat1}{HTML}{F7FBFF}  
\definecolor{heat2}{HTML}{DEEBF7}  
\definecolor{heat3}{HTML}{C6DBEF}  
\definecolor{heat4}{HTML}{9ECAE1}  
\definecolor{heat5}{HTML}{6BAED6}  
\definecolor{heat6}{HTML}{4292C6}  

\title{CurvDelta: Low-Curvature Delta Modeling \\For Surgical Learning and Unlearning}

\author{%
    Zeguan Xiao$^{1}$\thanks{Equal contribution.},
    Zhuocheng Liang$^{1}$\footnotemark[1],
    Yun Chen$^{1}$,
    Yong Wang$^{2}$,
    Guanhua Chen$^{3}$,
    Pinyan Lu$^{1}$ \\
    $^1$Shanghai University of Finance and Economics, $^2$Alibaba Group \\
    $^3$Southern University of Science and Technology
}

\iclrfinalcopy 
\begin{document}

\maketitle
\lhead{Preprint.}

\begin{abstract}
A central challenge in adapting large language models (LLMs) is capability preservation: an adaptation can achieve its target while degrading general capabilities. We introduce \mname, a scalable framework that incorporates the preservation constraint into the parameterization of the model update. By analyzing the local geometry of the preservation constraint around the pre-adaptation model, \mname constructs a low-curvature space in which parameter updates leave the preservation loss nearly invariant. We make its construction practical for LLMs through Kronecker-factored approximate curvature (K-FAC) and a factorized projection. We also introduce \mnamelora, a low-rank variant within the same preservation geometry, designed for more efficient adaptation. Both methods apply to preservation-constrained adaptation tasks such as capability learning and selective unlearning. We evaluate our methods on mathematical reasoning, code generation, and knowledge unlearning. For mathematical reasoning and code generation, our approaches achieve competitive downstream performance while better preserving general capabilities, offering a favorable acquisition--retention trade-off relative to the baselines. For knowledge unlearning, our main results on TOFU and MUSE show a better balance between targeted forgetting and knowledge retention than the baselines, with retention performance close to the from-scratch retraining reference.\footnote{Project page: \url{https://xiaozi-max.github.io/CurvDelta/}}
\end{abstract}

\section{Introduction}

Large language models (LLMs) need to evolve after pretraining. New tasks and domains call for additional capabilities, while privacy, safety, and data-governance requirements can necessitate selective unlearning. Retraining from scratch for every adaptation query is costly, motivating fine-tuning and parameter-efficient methods for capability learning and selective unlearning \citep{hu2022lora,cao2015towards,maini2024tofu,shi2024muse}. Although learning and unlearning change behavior in different ways, both require the update to be selective: progress on the target objective should preserve unrelated knowledge and capabilities.

To preserve existing capabilities while allowing the changes needed for adaptation, prior work has introduced different approaches to limiting unwanted changes. Replay and regularization incorporate preservation into the training objective \citep{rolnick2019experience,kirkpatrick2017overcoming}. Another line of work constrains the space in which adaptation may move, either by projecting task gradients \citep{chaudhry2019agem,ikram2026crispedit} or by heuristically restricting adaptation to orthogonal or null subspaces of parameters or representations \citep{fang2025alphaedit,xiong2026oplora,ju2026nullspace,liang2024inflora,luo2026keeplora}. Despite improvements, these methods either do not guarantee that the parameter update satisfies the intended constraint or impose strong constraints in parameter or representation space that are only indirectly tied to capability preservation. These limitations can lead to an unfavorable adaptation--preservation trade-off.

In this paper, we adopt a first-principles formulation of preservation-constrained adaptation. We seek a model delta that minimizes the task loss subject to a bound on the preservation loss, which measures changes in the model's behavior on a designated preservation set.
We therefore ask:
\begin{tcolorbox}[before skip=2mm, after skip=0.0cm, boxsep=0.0cm, middle=0.0cm, top=0.05cm, bottom=0.05cm, boxrule=0.6pt]
\begin{center}
    \textit{\textbf{(Q)} How can we directly incorporate the preservation constraint into the model delta parameterization while retaining the flexibility needed for adaptation?}
\end{center}
\end{tcolorbox}
\vspace*{2mm}

\begin{figure}[!t]
  \centering
  \includegraphics[width=\linewidth]{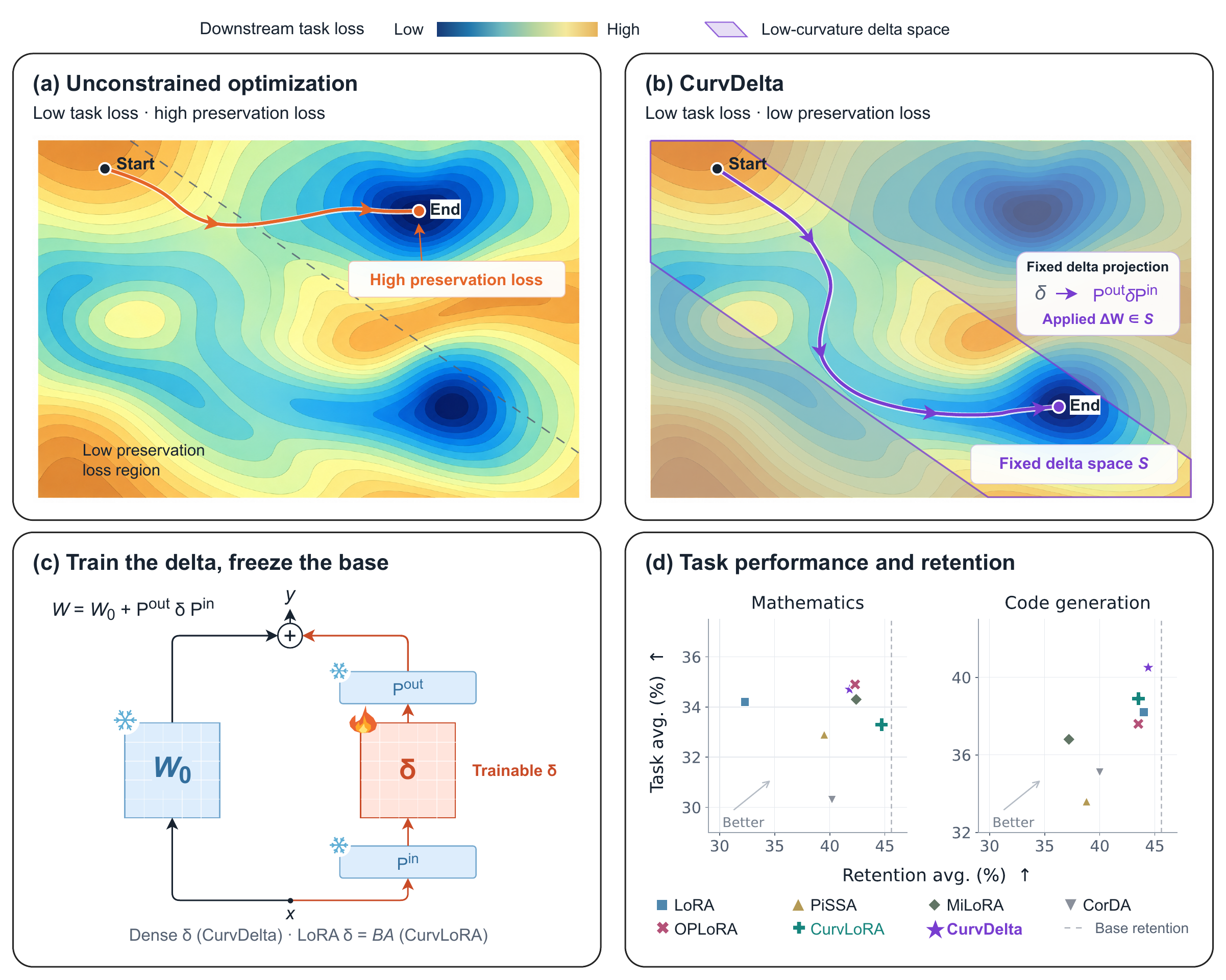}
  \vspace{-1.5mm}
  \caption{\textbf{Overview of \mname.}
  \textbf{(a)} Unconstrained optimization reaches low task loss but high preservation loss.
  \textbf{(b)} \mname constructs a low-curvature delta space. The applied delta remains in this space throughout optimization, reaching a different solution with comparable task loss.
  \textbf{(c)} The base weights and projections remain frozen. The trainable delta is a dense matrix in \mname or LoRA in \mnamelora.
  \textbf{(d)} Performance comparison. The dashed line marks the base model's retention average. Our methods achieve the best trade-off between task performance and capability retention.}
  \vspace{-8mm}
  \label{fig:overview}
\end{figure}

To address this question, we introduce \mname, a framework for low-curvature delta modeling for surgical learning and unlearning (Figure~\ref{fig:overview}). \mname learns a dense delta within a fixed low-curvature space. We also introduce \mnamelora, a low-rank variant that uses the same preservation geometry for more efficient adaptation. The method is built around the following three core ideas.

\noindent\textbf{1. Preservation through a task-independent delta parameterization.}
\mname derives a low-curvature delta space from the local geometry of the preservation constraint around the pre-adaptation model. It constrains a trainable delta into this fixed space before applying it, keeping the realized update constrained throughout optimization. Because the space depends only on the base model and preservation set, the same preservation geometry can be reused across task sets, task objectives, and delta parameterizations.

\noindent\textbf{2. Scalable delta space construction for dense and low-rank adaptation.}
To make low-curvature delta space construction tractable at LLM scale, \mname approximates preservation curvature with a block-diagonal Kronecker-factored approximate curvature (K-FAC) \citep{martens2015optimizing}. It further applies factorwise spectral truncation to obtain a separable two-sided projection that can be stored and applied efficiently using compact eigenvector bases. The same construction supports both dense adaptation in \mname and low-rank adaptation in \mnamelora.

\noindent\textbf{3. An explicit preservation guarantee from curvature and displacement.}
We show that preservation is jointly controlled by the curvature of the retained delta space and the magnitude of the realized update. Specifically, we derive an upper bound on the local preservation loss in which curvature determines the cost per unit displacement, while the delta norm determines the total displacement. Under the local quadratic and block-diagonal K-FAC models, this bound provides a sufficient condition for both \mname and \mnamelora to satisfy the preservation constraint.

\noindent\textbf{Experimental results.} We evaluate our framework on capability learning and selective unlearning. For capability learning, we study mathematical reasoning and code generation together with retention of general knowledge. Our approach achieves competitive downstream performance while reducing degradation in general capabilities, yielding a favorable acquisition--retention trade-off relative to the evaluated baselines. For selective unlearning, our main results on TOFU and MUSE show a better balance between targeted forgetting and knowledge retention than the evaluated baselines. On TOFU, our approach suppresses targeted knowledge while maintaining utility comparable to the retraining reference. On MUSE, it substantially reduces targeted memorization while retaining most of the retrained model's performance.

\section{Problem Formulation: Preservation-Constrained Adaptation}
\label{sec:prelim}

Let $\bm{\theta}_0$ be the initial parameter vector of a language model. We write $\bm{\theta}=\bm{\theta}_0+\Delta\bm{\theta}$ for the adapted parameters, with $\Delta\bm{\theta}$ the \emph{update}, and $p_{\bm{\theta}}$ for the distribution induced by $\bm{\theta}$.
Preservation-constrained adaptation seeks an update $\Delta\bm{\theta}$ that achieves a specified behavior change while leaving unrelated behaviors intact.
Each adaptation can be described by a \emph{task set} $\mathcal{D}_{\mathrm{task}}$, which encodes the desired behavior, and a \emph{preservation set} $\mathcal{D}_{\mathrm{pres}}$, which marks behavior that should remain unchanged. For capability learning, $\mathcal{D}_{\mathrm{task}}$ contains data from a new task; for selective unlearning, it contains examples of the behavior to be removed. $\mathcal{D}_{\mathrm{pres}}$ may comprise pretraining data or a representative proxy dataset.

Let $\mathcal{L}_{\mathrm{task}}(\bm{\theta})$ denote the task loss on $\mathcal{D}_{\mathrm{task}}$. We define the preservation loss as $\mathcal{L}_{\mathrm{pres}}(\bm{\theta})=\mathbb{E}_{x\sim\mathcal{D}_{\mathrm{pres}}}[d(\bm{z}_{\bm{\theta}}(x),\bm{z}_{\bm{\theta}_0}(x))]$, where $\bm{z}_{\bm{\theta}}(x)$ denotes the model output vector and $d$ measures its deviation from the original output.
We then formulate preservation-constrained adaptation as the following constrained optimization problem:
\begin{equation}
  \min_{\bm{\theta}}
  \mathcal{L}_{\mathrm{task}}(\bm{\theta})
  \quad
  \mathrm{s.t.}
  \quad
  \mathcal{L}_{\mathrm{pres}}(\bm{\theta})
  \le \varepsilon,
  \label{eq:prelim_constrained}
\end{equation}
where $\varepsilon>0$ specifies the tolerated preservation drift. This formulation covers both acquiring a new capability and removing a targeted behavior, with only the choice of $\mathcal{L}_{\mathrm{task}}$ distinguishing the two.

A standard way to optimize this problem is to minimize the Lagrangian $\mathcal{L}_{\mathrm{task}}(\bm{\theta})+\lambda\,\mathcal{L}_{\mathrm{pres}}(\bm{\theta})$, with $\lambda\geq0$ controlling the task--preservation trade-off. This relaxation has two limitations. First, $\lambda$ does not translate into the tolerance $\varepsilon$ of Eq.~\eqref{eq:prelim_constrained}: the drift is known only after training. Second, the Lagrangian is itself a multi-task objective, and gradient conflict makes the two objectives difficult to satisfy jointly \citep{yu2020gradient,liu2021conflict,chen2020just}, especially in unlearning \citep{xiao-etal-2026-modeling}.

\section{\mname: Low-Curvature Delta Modeling}
\label{sec:method}

We now present \mname, which turns the preservation constraint in Problem~\eqref{eq:prelim_constrained} into a restriction on the model update $\Delta\bm{\theta}$. The key idea is to optimize the realized delta within a low-curvature subspace of $\mathcal{L}_{\mathrm{pres}}$ at the original parameters $\bm{\theta}_0$.

\subsection{Local Curvature of the Preservation Loss}
\label{sec:method_taylor}

To justify restricting $\Delta\bm{\theta}$ to low-curvature directions, we first relate curvature to the preservation cost induced by an update. Since $\mathcal{L}_{\mathrm{pres}}$ measures the output discrepancy from the original model, we analyze its local variation around $\bm{\theta}_0$ under the following assumption on $d$.

\begin{assumption}
\label{as:method_discrepancy}
The map $\bm{\theta}\mapsto \bm{z}_{\bm{\theta}}(x)$ is twice continuously differentiable, and the discrepancy $d$ satisfies $d(\bm{v},\bm{v})=0$, $\nabla_{\bm{u}}d(\bm{u},\bm{v})|_{\bm{u}=\bm{v}}=0$, and $\nabla_{\bm{u}}^2d(\bm{u},\bm{v})|_{\bm{u}=\bm{v}}\succeq0$ for every model output $\bm{v}$.
\end{assumption}

These conditions are mild: a twice-differentiable nonnegative $d$ with $d(\bm{v},\bm{v})=0$ is minimized at $\bm{u}=\bm{v}$, so the remaining two conditions follow. Bregman divergences provide a broad family of such discrepancies \citep{bregman1967relaxation}; in particular, log-sum-exp on logits gives $d(\bm{z}_{\bm{\theta}},\bm{z}_{\bm{\theta}_0})=D_{\mathrm{KL}}(p_{\bm{\theta}_0}\|p_{\bm{\theta}})$, which we use throughout. Appendix~\ref{app:bregman} gives the derivation.

Under Assumption~\ref{as:method_discrepancy}, the second-order Taylor expansion around $\bm{\theta}_0$ yields the approximation:
\begin{equation}
  \mathcal{L}_{\mathrm{pres}}
  (\bm{\theta}_0+\Delta\bm{\theta})
  \approx
  \frac{1}{2}\Delta\bm{\theta}^\top
  \bm{H}_{\mathrm{pres}}
  \Delta\bm{\theta}.
  \label{eq:method_preservation_quadratic}
\end{equation}
where $\bm{H}_{\mathrm{pres}}:=\nabla_{\bm{\theta}}^2\mathcal{L}_{\mathrm{pres}}(\bm{\theta}_0)$.
By the second-order chain rule, $\bm{H}_{\mathrm{pres}}$ consists of the generalized Gauss--Newton (GGN) term below and a residual of the form $\mathbb{E}_x[\sum_k(\partial_{z_k}d)\nabla_{\bm{\theta}}^2z_{\bm{\theta},k}(x)]$. At $\bm{\theta}_0$, the two arguments of $d$ coincide, and Assumption~\ref{as:method_discrepancy} gives $\partial_{z_k}d=0$ for every $k$. The residual therefore vanishes, yielding
\begin{equation}
  \bm{G}_{\mathrm{pres}}
  :=
  \mathbb{E}_{x\sim\mathcal{D}_{\mathrm{pres}}}
  \!\left[
    \bm{J}_0(x)^\top \bm{H}_z(x)\bm{J}_0(x)
  \right]
  =\bm{H}_{\mathrm{pres}}
  \succeq \bm{0} ,
  \label{eq:method_preservation_gnh}
\end{equation}
Here, $\bm{J}_0(x)$ and $\bm{H}_z(x)$ are respectively the parameter Jacobian of the outputs and the output-space Hessian of $d$, both evaluated at $\bm{\theta}_0$. Since $\bm{H}_z(x)\succeq\bm{0}$, the resulting GGN matrix is positive semidefinite. Appendix~\ref{app:gnh} gives the full derivation.

Substituting the quadratic approximation into the constraint $\mathcal{L}_{\mathrm{pres}}\le\varepsilon$ gives
\begin{equation}
  \min_{\Delta\bm{\theta}}
  \mathcal{L}_{\mathrm{task}}
  (\bm{\theta}_0+\Delta\bm{\theta})
  \quad
  \mathrm{s.t.}
  \quad
  \Delta\bm{\theta}^\top \bm{G}_{\mathrm{pres}}\Delta\bm{\theta} \le 2\varepsilon.
  \label{eq:method_gnh_constrained}
\end{equation}
Let $(\lambda_i,\bm{u}_i)$ denote the eigenpairs of $\bm{G}_{\mathrm{pres}}$, and write $\Delta\bm{\theta}=\sum_i\alpha_i\bm{u}_i$. Then the constraint becomes $\sum_i\lambda_i\alpha_i^2\le2\varepsilon$, whereas $\|\Delta\bm{\theta}\|_2^2=\sum_i\alpha_i^2$. Thus, for a fixed update norm, concentrating the update on eigenvectors with smaller eigenvalues incurs less preservation loss. We therefore construct the delta space from the trailing eigenspace of $\bm{G}_{\mathrm{pres}}$, which spans its low-curvature subspace.
Appendix~\ref{app:preservation_spectrum} gives the full derivation.

\subsection{Constructing the Low-Curvature Delta Space}
\label{sec:method_space}

Computing the trailing eigenspace of $\bm{G}_{\mathrm{pres}}$ is intractable at LLM scale. We therefore approximate $\bm{G}_{\mathrm{pres}}$ with block-diagonal K-FAC \citep{martens2015optimizing}, using one block for each adapted linear weight. For $\bm{W}_l\in\mathbb{R}^{d_{\mathrm{out}}\times d_{\mathrm{in}}}$, the corresponding block is
\begin{equation}
  \bm{G}_{\mathrm{pres}}^{(l)}
  \approx \bm{K}_l^{\mathrm{in}}\otimes \bm{K}_l^{\mathrm{out}},
  \qquad
  \bm{K}_l^{\mathrm{in}}=\mathbb{E}[\bm{a}_{l-1}\bm{a}_{l-1}^\top],
  \quad
  \bm{K}_l^{\mathrm{out}}=\mathbb{E}[\bm{g}_l\bm{g}_l^\top].
  \label{eq:method_kfac}
\end{equation}
Here, $\bm{a}_{l-1}$ is the input activation and $\bm{g}_l$ is the layer output gradient. We estimate the factors once at $\bm{\theta}_0$ on $\mathcal{D}_{\mathrm{pres}}$ using the Monte Carlo Fisher estimator in Appendix~\ref{app:mc_fisher}.

For clarity, we omit the layer index below. Let $(\lambda_i^{\mathrm{out}},\bm{u}_i^{\mathrm{out}})$ and $(\lambda_j^{\mathrm{in}},\bm{u}_j^{\mathrm{in}})$ denote the factor eigenpairs, with eigenvalues in descending order. By the Kronecker eigenstructure, the matrix direction $\bm{u}_i^{\mathrm{out}}(\bm{u}_j^{\mathrm{in}})^\top$ has curvature $\lambda_i^{\mathrm{out}}\lambda_j^{\mathrm{in}}$ under the K-FAC block. We construct a low-curvature delta space by retaining directions formed from trailing eigenvectors of both factors.

\paragraph{Factorwise spectral truncation.}
For $\gamma\in(0,1)$, let $r_{\mathrm{out}}$ and $r_{\mathrm{in}}$ be the smallest ranks at which the cumulative spectral energy reaches $\gamma$ for the respective factors. Let $\bm{U}_h^{\mathrm{out}}$ and $\bm{U}_h^{\mathrm{in}}$ contain these leading eigenvectors. The orthogonal projectors onto the complementary trailing eigenspaces are
\begin{equation}
  \bm{P}^{\mathrm{out}}
  =\bm{I}-\bm{U}_h^{\mathrm{out}}(\bm{U}_h^{\mathrm{out}})^\top,
  \qquad
  \bm{P}^{\mathrm{in}}
  =\bm{I}-\bm{U}_h^{\mathrm{in}}(\bm{U}_h^{\mathrm{in}})^\top.
  \label{eq:method_delta_projectors}
\end{equation}
Given a trainable pre-projection delta $\bm{\delta}(\bm{\phi})$ with the same shape as $\bm{W}$, we define the applied delta as
\begin{equation}
  \Delta\bm{W}
  :=\bm{P}^{\mathrm{out}}\,\bm{\delta}(\bm{\phi})\,\bm{P}^{\mathrm{in}}.
  \label{eq:method_delta_projection}
\end{equation}
This projection retains exactly the span of $\bm{u}_i^{\mathrm{out}}(\bm{u}_j^{\mathrm{in}})^\top$ with $i>r_{\mathrm{out}}$ and $j>r_{\mathrm{in}}$. The resulting delta space has dimension $(d_{\mathrm{out}}-r_{\mathrm{out}})(d_{\mathrm{in}}-r_{\mathrm{in}})$. Appendix~\ref{app:low_curvature_spectrum} gives the full derivation.

\paragraph{Curvature of the retained space.}
The largest eigenvalue in the retained space is the product of the largest retained factor eigenvalues. Expanding $\Delta\bm{W}$ in the corresponding orthonormal basis therefore yields
\begin{equation*}
  \begin{gathered}
    \kappa:=\lambda_{r_{\mathrm{out}}+1}^{\mathrm{out}}
      \lambda_{r_{\mathrm{in}}+1}^{\mathrm{in}},\\
    \operatorname{vec}(\Delta\bm{W})^\top
    (\bm{K}^{\mathrm{in}}\otimes\bm{K}^{\mathrm{out}})
    \operatorname{vec}(\Delta\bm{W})
    \le\kappa\|\Delta\bm{W}\|_F^2,
  \end{gathered}
\end{equation*}
Thus, under the K-FAC model, the layer's contribution to the quadratic preservation loss is bounded by $\tfrac12\kappa\|\Delta\bm{W}\|_F^2$ (Appendix~\ref{app:low_curvature_control}). As $\gamma$ increases, both $\kappa$ and the dimension of the retained space are nonincreasing, so the threshold jointly controls the curvature bound and the available update space.

Our factorwise construction yields a subspace of the full K-FAC eigenspace with eigenvalues at most $\kappa$. Unlike product-spectrum thresholding \citep{ikram2026crispedit}, which applies a non-separable elementwise mask across the full Kronecker eigenbasis, factorwise truncation produces the separable two-sided projection in Eq.~\eqref{eq:method_delta_projection}. This substantially reduces storage and computational costs by caching only the leading factor eigenvectors, and allows applying the projection directly to low-rank factors without materializing a dense matrix, yielding \mnamelora. While it may discard mixed directions where a small eigenvalue in one factor offsets a large eigenvalue in the other, empirical evidence of abundant flat directions~\citep{sagun2017empirical,pmlr-v97-ghorbani19b} and low-dimensional optimization~\citep{li2018measuring,aghajanyan2021intrinsic} suggests that the retained space still provides sufficient capacity for adaptation (Appendix~\ref{app:product_threshold}).

\subsection{Delta Parameterizations and Optimization}
\label{sec:method_algorithm}

Having constructed the fixed low-curvature update space in Section~\ref{sec:method_space}, we now turn to representing and optimizing task-specific updates within it. The projected form supports two complementary parameterizations: a dense delta that spans the entire retained space and a low-rank LoRA delta that reduces the number of trainable parameters, yielding \mname and \mnamelora, respectively.

\begin{algorithm}[ht]
\caption{Low-Curvature Delta Modeling}
\label{alg:curvdelta}
\begin{algorithmic}[1]
\REQUIRE Initial parameters $\bm{\theta}_0$; task and preservation sets $\mathcal{D}_{\mathrm{task}}$, $\mathcal{D}_{\mathrm{pres}}$; task loss $\mathcal{L}_{\mathrm{task}}$.
\REQUIRE Target weights $\mathcal{T}$; threshold $\gamma$; delta parameterization $\bm{\delta}_l(\bm{\phi}_l)$; learning rates $\{\eta_t\}_{t=1}^T$.
\ENSURE Adapted parameters $\bm{\theta}_0+\Delta\bm{\theta}$.
\STATE Estimate the K-FAC factors at $\bm{\theta}_0$ on $\mathcal{D}_{\mathrm{pres}}$ (Appendix~\ref{app:mc_fisher}).
\STATE Cache the leading factor eigenbases selected by $\gamma$.
\STATE Initialize $\bm{\phi}_l$; keep $\bm{\theta}_0$ fixed.
\FOR{$t=1$ to $T$}
  \STATE For each $l\in\mathcal{T}$, form the projected delta $\Delta\bm{W}_l=\bm{P}_l^{\mathrm{out}}\bm{\delta}_l(\bm{\phi}_l)\bm{P}_l^{\mathrm{in}}$.
  \STATE Take an optimizer step on $\{\bm{\phi}_l\}_{l\in\mathcal{T}}$ using the task loss $\mathcal{L}_{\mathrm{task}}(\bm{\theta}_0+\Delta\bm{\theta})$.
\ENDFOR
\STATE Recompute $\Delta\bm{\theta}=\{\Delta\bm{W}_l\}_{l\in\mathcal{T}}$ from the final delta parameters and merge into $\bm{\theta}_0$.
\STATE \textbf{return} $\bm{\theta}_0+\Delta\bm{\theta}$.
\end{algorithmic}
\end{algorithm}

\paragraph{Dense parameterization.}
\mname parameterizes the delta as a dense matrix $\bm{\delta}_l\in\mathbb{R}^{d_{\mathrm{out}}\times d_{\mathrm{in}}}$ initialized at zero. Under the projection in Eq.~\eqref{eq:method_delta_projection}, this parameterization spans the entire low-curvature space. It has $d_{\mathrm{out}}d_{\mathrm{in}}$ trainable parameters per weight and $(d_{\mathrm{out}}-r_{\mathrm{out}})(d_{\mathrm{in}}-r_{\mathrm{in}})$ effective degrees of freedom after projection.

\paragraph{Low-rank parameterization.}
\mnamelora reduces the number of trainable parameters through a LoRA factorization~\citep{hu2022lora},
\begin{equation*}
  \bm{\delta}_l=\bm{B}_l\bm{A}_l,
  \qquad
  \Delta\bm{W}_l
  =(\bm{P}_l^{\mathrm{out}}\bm{B}_l)
   (\bm{A}_l\bm{P}_l^{\mathrm{in}}),
\end{equation*}
where $\bm{B}_l\in\mathbb{R}^{d_{\mathrm{out}}\times r}$ and $\bm{A}_l\in\mathbb{R}^{r\times d_{\mathrm{in}}}$, with $\bm{B}_l$ initialized at zero. This parameterization has $r(d_{\mathrm{out}}+d_{\mathrm{in}})$ trainable parameters per weight and restricts the applied delta to rank at most $r$ within the same low-curvature space. The factorized form avoids materializing a dense weight matrix.

Both variants apply the projection implicitly through the cached leading eigenvector bases. The corresponding computations and their time and memory complexity are detailed in Section~\ref{sec:exp_analysis} and Appendix~\ref{app:complexity}.

\paragraph{Optimization.}
The preservation geometry is estimated once before training, and the resulting projectors and base parameters remain fixed. At each step, layer $l$ uses $\bm{W}_{l,0}+\Delta\bm{W}_l$, where $\Delta\bm{W}_l$ is the actual update applied to $\bm{W}_{l,0}$. We minimize $\mathcal{L}_{\mathrm{task}}(\bm{\theta}_0+\Delta\bm{\theta})$ over the delta parameters---$\bm{\delta}_l$ for \mname and $(\bm{B}_l,\bm{A}_l)$ for \mnamelora---by differentiating through the projection. Because the projection is built into the parameterization, the realized update $\Delta\bm{W}_l$ remains in the prescribed space throughout optimization, regardless of the optimizer. For dense updates, ordinary SGD from zero is equivalent to gradient projection onto the same fixed space; under adaptive optimizers such as Adam, however, gradient projection \citep{ikram2026crispedit} rescales coordinates by unequal second-moment factors, moving the update outside that space (Algorithm~\ref{alg:curvdelta}).

\subsection{Preservation Guarantee}
\label{sec:method_guarantee}

Under the block-diagonal K-FAC model, the quadratic preservation loss decomposes over the adapted weights. Summing the layerwise bounds from Section~\ref{sec:method_space} therefore gives the following guarantee.

\begin{theorem}[$\varepsilon$-feasible preservation]
\label{thm:method_epsilon_feasible}
Under the local quadratic model of Section~\ref{sec:method_taylor} and the block-diagonal K-FAC model of Assumption~\ref{app:as_blockdiag}, let $\Delta\bm{\theta}$ be supported on $\mathcal{T}$ with layer deltas given by Eq.~\eqref{eq:method_delta_projection}. Then
\begin{equation}
  \mathcal{L}_{\mathrm{pres}}(\bm{\theta}_0+\Delta\bm{\theta})
  \approx \frac{1}{2}\Delta\bm{\theta}^\top
  \bm{G}_{\mathrm{pres}}\Delta\bm{\theta}
  \le \frac{1}{2}\sum_{l\in\mathcal{T}}
  \kappa_l\|\Delta\bm{W}_l\|_F^2,
  \label{eq:method_drift_bound}
\end{equation}
where $\kappa_l$ is the curvature coefficient for layer $l$ from Section~\ref{sec:method_space}. The update is feasible under the local quadratic model whenever $\sum_{l\in\mathcal{T}}\kappa_l\|\Delta\bm{W}_l\|_F^2\le2\varepsilon$.
\end{theorem}

The proof applies Proposition~\ref{app:prop_low_curvature} to each block (Appendix~\ref{app:thm_epsilon}). The geometry fixes $\kappa_l$, while the delta norms depend on the optimization trajectory, including the learning-rate schedule and number of training steps.

\section{Experiments}
\label{sec:experiments}

We summarize the setup and main results here, and provide the experimental details in Appendix~\ref{app:experimental_details}.

\subsection{Capability Learning}
\label{sec:exp_learning}

\subsubsection{Experimental Setup}
\label{sec:exp_learning_setup}

\paragraph{Tasks, datasets, and evaluation.} We study capability acquisition in mathematical reasoning and code generation. For mathematical reasoning, we fine-tune on 100K examples sampled from MetaMathQA \citep{yu2024metamath} and evaluate on GSM8K \citep{cobbe2021gsm8k} and MATH \citep{hendrycks2021math}. For code generation, we fine-tune on 104K examples sampled from CodeFeedback \citep{zheng2024opencodeinterpreter} and evaluate on HumanEval \citep{chen2021humaneval} and MBPP \citep{austin2021mbpp}.
To measure knowledge retention, we evaluate on TriviaQA \citep{joshi2017triviaqa}, NQ-Open \citep{kwiatkowski2019naturalquestions}, and MMLU \citep{hendrycks2021measuring}.
For the \emph{preservation set} $\mathcal{D}_{\mathrm{pres}}$, we use 256 examples sampled from the FineFineWeb corpus \citep{map2024finefineweb}.

\paragraph{Models.} We evaluate both tasks using the pretrained base checkpoints of Gemma-3-1B \citep{gemmateam2025gemma3} and Llama-2-7B \citep{touvron2023llama2}. For PEFT methods and \mnamelora, we set the rank to 128 for Gemma-3-1B and 32 for Llama-2-7B, holding it fixed across tasks.

\paragraph{Baselines.} We include standard fine-tuning (FT) as an unconstrained adaptation reference and experience replay (Replay) \citep{rolnick2019experience} as a data-based baseline. Replay mixes task examples with examples sampled from $\mathcal{D}_{\mathrm{pres}}$ during fine-tuning. We additionally include CrispEdit \citep{ikram2026crispedit} as a curvature-based gradient-projection baseline and OPLoRA \citep{xiong2026oplora} as an orthogonal delta-projection baseline. We further compare against LoRA \citep{hu2022lora}, PiSSA \citep{meng2024pissa}, MiLoRA \citep{wang2025milora}, and the knowledge-preserved version of CorDA \citep{yang2024corda}.

\subsubsection{Main Results}
\label{sec:exp_learning_results}

Tables~\ref{tab:main_learning_math} and~\ref{tab:main_learning_code} compare the acquisition--retention trade-off in mathematical reasoning and code generation. Full fine-tuning provides strong capability acquisition but consistently reduces general-knowledge retention, whereas \mname keeps downstream performance competitive while preserving substantially more general knowledge across both domains. This pattern is strongest for code generation: \mnamedense attains the strongest joint averages on both backbones, with downstream performance close to or above full fine-tuning and the highest average retention among the reported adaptation methods. On mathematical reasoning, \mnamedense remains close to full fine-tuning in downstream performance while substantially narrowing the retention gap. The low-rank trade-off varies by backbone: \mnamelora achieves the highest average retention among adaptation methods on Llama, though with lower downstream performance than \mnamedense, while \mnamedense provides the stronger joint result on Gemma. Overall, restricting updates to the low-curvature space shifts the acquisition--retention trade-off toward better general-knowledge retention; the dense variant delivers stronger joint results in most task--backbone pairs, while \mnamelora remains competitive with fewer trainable parameters.

\begin{table}[t]
\centering
\scriptsize
\setlength{\tabcolsep}{3.2pt}
\caption{\textbf{Capability learning on mathematical reasoning.} Within each backbone, the best and second-best results among adaptation methods are shown in bold and underlined, respectively.}
\label{tab:main_learning_math}
\begin{tabularx}{\textwidth}{l *{8}{Y}}
\toprule
\multirow{2}{*}{\textbf{Method}} & \multicolumn{3}{c}{\textbf{Downstream tasks} (\%)} & \multicolumn{4}{c}{\textbf{Retention tasks} (\%)} & \multirow{2}{*}{\textbf{ALL}} \\
\cmidrule(lr){2-4}\cmidrule(lr){5-8}
& \textbf{GSM8K} & \textbf{MATH} & \textbf{Avg.} & \textbf{TriviaQA} & \textbf{NQ-Open} & \textbf{MMLU} & \textbf{Avg.} & \\
\midrule
\rowcolor{basebg}
Gemma-3-1B & 0.0 & 0.0 & \cellcolor{avgbg}0.0 & 35.7 & 10.7 & 26.1 & \cellcolor{avgbg}24.2 & \cellcolor{avgbg}14.5 \\
$\hookrightarrow$\,FT & \textbf{36.9} & 5.7 & \cellcolor{avgbg}\underline{21.3} & 22.5 & 6.7 & 25.2 & \cellcolor{avgbg}18.1 & \cellcolor{avgbg}19.4 \\
$\hookrightarrow$\,Replay & \underline{36.8} & \textbf{6.3} & \cellcolor{avgbg}\textbf{21.6} & 13.9 & 4.9 & 24.4 & \cellcolor{avgbg}14.4 & \cellcolor{avgbg}17.3 \\
$\hookrightarrow$\,CrispEdit & 32.8 & 5.2 & \cellcolor{avgbg}19.0 & \textbf{34.0} & \underline{10.1} & \underline{26.1} & \cellcolor{avgbg}\textbf{23.4} & \cellcolor{avgbg}\underline{21.6} \\
$\hookrightarrow$\,LoRA & 34.8 & \underline{5.9} & \cellcolor{avgbg}20.4 & 15.8 & 5.3 & 25.6 & \cellcolor{avgbg}15.6 & \cellcolor{avgbg}17.5 \\
$\hookrightarrow$\,PiSSA & 35.4 & 5.4 & \cellcolor{avgbg}20.4 & 11.6 & 3.3 & 26.0 & \cellcolor{avgbg}13.6 & \cellcolor{avgbg}16.3 \\
$\hookrightarrow$\,MiLoRA & 36.0 & \underline{5.9} & \cellcolor{avgbg}21.0 & 15.4 & 5.5 & 24.8 & \cellcolor{avgbg}15.2 & \cellcolor{avgbg}17.5 \\
$\hookrightarrow$\,CorDA & 30.0 & 5.1 & \cellcolor{avgbg}17.6 & 25.7 & 8.2 & \textbf{26.2} & \cellcolor{avgbg}20.0 & \cellcolor{avgbg}19.0 \\
$\hookrightarrow$\,OPLoRA & 33.7 & 5.1 & \cellcolor{avgbg}19.4 & 27.3 & 8.6 & 26.0 & \cellcolor{avgbg}20.6 & \cellcolor{avgbg}20.1 \\
\rowcolor{methodbg}
$\hookrightarrow$\,\mnamelora & 33.7 & 4.7 & \cellcolor{methodbg}19.2 & 30.1 & 8.7 & 25.0 & \cellcolor{methodbg}21.3 & \cellcolor{methodbg}20.4 \\
\rowcolor{methodbg}
$\hookrightarrow$\,\mnamedense & \textbf{36.9} & 5.5 & \cellcolor{methodbg}21.2 & \underline{32.5} & \textbf{10.2} & \underline{26.1} & \cellcolor{methodbg}\underline{22.9} & \cellcolor{methodbg}\textbf{22.2} \\
\midrule
\rowcolor{basebg}
Llama-2-7B & 0.0 & 0.0 & \cellcolor{avgbg}0.0 & 64.4 & 26.6 & 45.9 & \cellcolor{avgbg}45.6 & \cellcolor{avgbg}27.4 \\
$\hookrightarrow$\,FT & 59.3 & \textbf{12.6} & \cellcolor{avgbg}\underline{36.0} & 44.9 & 15.2 & 35.1 & \cellcolor{avgbg}31.7 & \cellcolor{avgbg}33.4 \\
$\hookrightarrow$\,Replay & \textbf{61.2} & \underline{12.0} & \cellcolor{avgbg}\textbf{36.6} & 59.8 & 24.4 & 43.4 & \cellcolor{avgbg}\underline{42.5} & \cellcolor{avgbg}\textbf{40.2} \\
$\hookrightarrow$\,CrispEdit & \underline{60.1} & 11.9 & \cellcolor{avgbg}\underline{36.0} & 51.0 & 19.2 & 40.6 & \cellcolor{avgbg}36.9 & \cellcolor{avgbg}36.6 \\
$\hookrightarrow$\,LoRA & 57.5 & 10.8 & \cellcolor{avgbg}34.2 & 47.0 & 16.6 & 33.3 & \cellcolor{avgbg}32.3 & \cellcolor{avgbg}33.0 \\
$\hookrightarrow$\,PiSSA & 55.8 & 9.9 & \cellcolor{avgbg}32.9 & 56.7 & 21.6 & 40.3 & \cellcolor{avgbg}39.5 & \cellcolor{avgbg}36.9 \\
$\hookrightarrow$\,MiLoRA & 58.0 & 10.5 & \cellcolor{avgbg}34.3 & 59.2 & 23.8 & \underline{44.3} & \cellcolor{avgbg}42.4 & \cellcolor{avgbg}39.2 \\
$\hookrightarrow$\,CorDA & 52.8 & 7.8 & \cellcolor{avgbg}30.3 & 57.0 & 21.7 & 41.8 & \cellcolor{avgbg}40.2 & \cellcolor{avgbg}36.2 \\
$\hookrightarrow$\,OPLoRA & 58.6 & 11.2 & \cellcolor{avgbg}34.9 & \underline{60.2} & 23.6 & 43.2 & \cellcolor{avgbg}42.3 & \cellcolor{avgbg}39.4 \\
\rowcolor{methodbg}
$\hookrightarrow$\,\mnamelora & 56.6 & 10.0 & \cellcolor{methodbg}33.3 & \textbf{62.9} & \textbf{26.1} & \textbf{45.1} & \cellcolor{methodbg}\textbf{44.7} & \cellcolor{methodbg}\underline{40.1} \\
\rowcolor{methodbg}
$\hookrightarrow$\,\mnamedense & 58.4 & 11.0 & \cellcolor{methodbg}34.7 & 58.7 & \underline{24.5} & 42.2 & \cellcolor{methodbg}41.8 & \cellcolor{methodbg}39.0 \\
\bottomrule
\end{tabularx}
\end{table}

\begin{table}[t]
\centering
\scriptsize
\setlength{\tabcolsep}{3.2pt}
\caption{\textbf{Capability learning on code generation.} Within each backbone, the best and second-best results among adaptation methods are shown in bold and underlined, respectively.}
\label{tab:main_learning_code}
\begin{tabularx}{\textwidth}{l *{8}{Y}}
\toprule
\multirow{2}{*}{\textbf{Method}} & \multicolumn{3}{c}{\textbf{Downstream tasks} (\%)} & \multicolumn{4}{c}{\textbf{Retention tasks} (\%)} & \multirow{2}{*}{\textbf{ALL}} \\
\cmidrule(lr){2-4}\cmidrule(lr){5-8}
& \textbf{HumanEval} & \textbf{MBPP} & \textbf{Avg.} & \textbf{TriviaQA} & \textbf{NQ-Open} & \textbf{MMLU} & \textbf{Avg.} & \\
\midrule
\rowcolor{basebg}
Gemma-3-1B & 1.2 & 13.0 & \cellcolor{avgbg}7.1 & 35.8 & 10.8 & 26.1 & \cellcolor{avgbg}24.2 & \cellcolor{avgbg}17.4 \\
$\hookrightarrow$\,FT & \textbf{20.7} & 19.3 & \cellcolor{avgbg}\textbf{20.0} & 29.2 & 10.1 & 25.4 & \cellcolor{avgbg}21.6 & \cellcolor{avgbg}20.9 \\
$\hookrightarrow$\,Replay & 16.5 & 21.7 & \cellcolor{avgbg}19.1 & 32.3 & 10.1 & \textbf{26.9} & \cellcolor{avgbg}23.1 & \cellcolor{avgbg}21.5 \\
$\hookrightarrow$\,CrispEdit & 15.9 & \textbf{22.5} & \cellcolor{avgbg}19.2 & \textbf{34.7} & \underline{10.5} & 25.7 & \cellcolor{avgbg}\underline{23.6} & \cellcolor{avgbg}\underline{21.9} \\
$\hookrightarrow$\,LoRA & 15.2 & 19.3 & \cellcolor{avgbg}17.3 & 32.4 & \textbf{10.8} & 26.2 & \cellcolor{avgbg}23.1 & \cellcolor{avgbg}20.8 \\
$\hookrightarrow$\,PiSSA & 17.1 & 20.9 & \cellcolor{avgbg}19.0 & 29.8 & 9.2 & 25.8 & \cellcolor{avgbg}21.6 & \cellcolor{avgbg}20.6 \\
$\hookrightarrow$\,MiLoRA & 14.6 & 19.3 & \cellcolor{avgbg}17.0 & 30.7 & 9.8 & 25.7 & \cellcolor{avgbg}22.1 & \cellcolor{avgbg}20.0 \\
$\hookrightarrow$\,CorDA & 15.2 & 18.3 & \cellcolor{avgbg}16.8 & 29.7 & 9.5 & 25.9 & \cellcolor{avgbg}21.7 & \cellcolor{avgbg}19.7 \\
$\hookrightarrow$\,OPLoRA & \underline{17.5} & 18.7 & \cellcolor{avgbg}18.1 & 32.5 & \textbf{10.8} & 25.9 & \cellcolor{avgbg}23.1 & \cellcolor{avgbg}21.1 \\
\rowcolor{methodbg}
$\hookrightarrow$\,\mnamelora & 16.5 & 20.6 & \cellcolor{methodbg}18.6 & 32.7 & 10.3 & \underline{26.3} & \cellcolor{methodbg}23.1 & \cellcolor{methodbg}21.3 \\
\rowcolor{methodbg}
$\hookrightarrow$\,\mnamedense & 17.1 & \underline{22.2} & \cellcolor{methodbg}\underline{19.7} & \underline{34.5} & \underline{10.5} & 26.1 & \cellcolor{methodbg}\textbf{23.7} & \cellcolor{methodbg}\textbf{22.1} \\
\midrule
\rowcolor{basebg}
Llama-2-7B & 1.8 & 29.9 & \cellcolor{avgbg}15.9 & 64.4 & 26.6 & 45.9 & \cellcolor{avgbg}45.6 & \cellcolor{avgbg}33.7 \\
$\hookrightarrow$\,FT & 37.2 & 39.7 & \cellcolor{avgbg}38.5 & 58.1 & 22.4 & 40.8 & \cellcolor{avgbg}40.4 & \cellcolor{avgbg}39.6 \\
$\hookrightarrow$\,Replay & \textbf{39.6} & 41.0 & \cellcolor{avgbg}\underline{40.3} & 58.3 & 22.3 & 41.8 & \cellcolor{avgbg}40.8 & \cellcolor{avgbg}40.6 \\
$\hookrightarrow$\,CrispEdit & \underline{37.8} & 40.7 & \cellcolor{avgbg}39.3 & 57.8 & 19.5 & 43.5 & \cellcolor{avgbg}40.3 & \cellcolor{avgbg}39.9 \\
$\hookrightarrow$\,LoRA & 34.1 & \textbf{42.3} & \cellcolor{avgbg}38.2 & \underline{62.7} & \textbf{25.7} & 43.5 & \cellcolor{avgbg}\underline{44.0} & \cellcolor{avgbg}\underline{41.7} \\
$\hookrightarrow$\,PiSSA & 31.1 & 36.0 & \cellcolor{avgbg}33.6 & 56.3 & 20.2 & 39.9 & \cellcolor{avgbg}38.8 & \cellcolor{avgbg}36.7 \\
$\hookrightarrow$\,MiLoRA & 35.4 & 38.1 & \cellcolor{avgbg}36.8 & 53.6 & 19.5 & 38.6 & \cellcolor{avgbg}37.2 & \cellcolor{avgbg}37.0 \\
$\hookrightarrow$\,CorDA & 29.9 & 40.2 & \cellcolor{avgbg}35.1 & 57.2 & 21.7 & 41.2 & \cellcolor{avgbg}40.0 & \cellcolor{avgbg}38.0 \\
$\hookrightarrow$\,OPLoRA & 35.4 & 39.7 & \cellcolor{avgbg}37.6 & 61.9 & 24.7 & \underline{43.9} & \cellcolor{avgbg}43.5 & \cellcolor{avgbg}41.1 \\
\rowcolor{methodbg}
$\hookrightarrow$\,\mnamelora & 36.0 & \underline{41.8} & \cellcolor{methodbg}38.9 & 62.4 & 24.8 & 43.3 & \cellcolor{methodbg}43.5 & \cellcolor{methodbg}\underline{41.7} \\
\rowcolor{methodbg}
$\hookrightarrow$\,\mnamedense & \textbf{39.6} & 41.3 & \cellcolor{methodbg}\textbf{40.5} & \textbf{63.1} & \underline{24.9} & \textbf{45.1} & \cellcolor{methodbg}\textbf{44.4} & \cellcolor{methodbg}\textbf{42.8} \\
\bottomrule
\end{tabularx}
\end{table}

\subsection{Selective Unlearning}
\label{sec:exp_unlearning}

\subsubsection{Experimental Setup}
\label{sec:exp_unlearning_setup}

\paragraph{Benchmarks and models.} We conduct experiments on two widely-used LLM unlearning benchmarks: TOFU \citep{maini2024tofu} and MUSE \citep{shi2024muse}. TOFU provides a controlled unlearning setting based on synthetic author-profile QA pairs, whereas MUSE evaluates verbatim and knowledge memorization on real-world text corpora.
For TOFU, we evaluate the Forget 5\% and Forget 10\% settings on three instruction-tuned backbones spanning 3B--8B scales: Llama-3.2-3B-Instruct \citep{grattafiori2024llama3}, Qwen3-4B-Instruct-2507 \citep{yang2025qwen3}, and Llama-3.1-8B-Instruct \citep{grattafiori2024llama3}. We omit the Forget 1\% setting because it is too weak to distinguish methods. For MUSE, we use the benchmark's Llama-2-7B target model on the Books corpus. Further details are provided in Appendix~\ref{app:unlearning_training_details}.

\paragraph{Baselines and objectives.} We compare \mnamelora and \mnamedense against five baselines: full fine-tuning (FT), CrispEdit \citep{ikram2026crispedit}, LoRA \citep{hu2022lora}, VILA \citep{kim2025improving}, and OPLoRA \citep{xiong2026oplora}. FT, LoRA, VILA, and OPLoRA represent full-parameter, standard parameter-efficient, unlearning-specific parameterizations, and orthogonal delta-projection adaptation, respectively, while CrispEdit is a curvature-based gradient-projection method. Each method is evaluated with three unlearning objectives: Gradient Difference (GradDiff) \citep{liu2022continual,maini2024tofu}, Negative Preference Optimization (NPO) \citep{zhang2024negative}, and SimNPO \citep{fan2025simplicity}. Appendices~\ref{app:objectives} and~\ref{app:unlearning_training_details} provide the objective definitions and optimization settings, respectively.

\paragraph{Evaluation.} For TOFU, we report three core metrics: Forget Quality (FQ), QA probability (FP), and model utility (MU). FQ is based on a KS-test $p$-value that checks whether the unlearned model and a model retrained from scratch without the forget data behave similarly on the forget examples. We report the base-10 logarithm of this $p$-value in the TOFU tables and denote the retrained model Retrain. For MUSE, we report forget verbatim memorization (VerbMem), forget knowledge memorization (KnowMem), and retain knowledge memorization (Retain KnowMem), all measured by ROUGE.

\subsubsection{Main Results}
\label{sec:exp_unlearning_results}

Table~\ref{tab:main_unlearning} reports the main results using SimNPO, with Llama-3.1-8B-Instruct on TOFU and Llama-2-7B on MUSE. Additional TOFU results cover two more backbones (Llama-3.2-3B-Instruct and Qwen3-4B-Instruct-2507) and two alternative objectives (GradDiff and NPO) in Appendix Tables~\ref{tab:app_tofu_llama32}--\ref{tab:app_tofu_qwen3}. MUSE results with these alternative objectives appear in Appendix Table~\ref{tab:app_muse_full_grid}.

\begin{table}[!t]
\centering
\scriptsize
\setlength{\tabcolsep}{3.2pt}
\caption{Unlearning under SimNPO. \mnamelora is the low-rank variant of \mname. Left: TOFU with Llama-3.1-8B-Instruct. Right: MUSE Books with Llama-2-7B.}
\label{tab:main_unlearning}
\begin{tabularx}{\textwidth}{l *{9}{Y}}
\toprule
& \multicolumn{6}{c}{\textbf{TOFU}} & \multicolumn{3}{c}{\textbf{MUSE}} \\
\cmidrule(lr){2-7}\cmidrule(lr){8-10}
\multirow{2}{*}{\textbf{Method}} & \multicolumn{3}{c}{\textbf{Forget 5\%}} & \multicolumn{3}{c}{\textbf{Forget 10\%}} & \multicolumn{2}{c}{\textbf{Forget}} & \mbox{\textbf{Retain}} \\
\cmidrule(lr){2-4}\cmidrule(lr){5-7}\cmidrule(lr){8-9}\cmidrule(lr){10-10}
 & \mbox{FQ $\uparrow$} & \mbox{FP $\downarrow$} & \mbox{MU $\uparrow$} & \mbox{FQ $\uparrow$} & \mbox{FP $\downarrow$} & \mbox{MU $\uparrow$} & \mbox{VerbMem $\downarrow$} & \mbox{KnowMem $\downarrow$} & \mbox{KnowMem $\uparrow$} \\
\midrule
\rowcolor{basebg}
Retrain & -- & 10.8 & 63.2 & -- & 10.6 & 64.6 & 14.2 & 29.5 & 68.6 \\
\rowcolor{basebg}
Original & $-11.8$ & 99.0 & 62.7 & $-26.8$ & 99.2 & 62.7 & 99.7 & 46.6 & 69.1 \\
$\hookrightarrow$\,FT & $-0.5$ & 13.0 & 54.6 & $-1.4$ & 14.4 & 38.9 & 23.3 & 27.6 & 46.8 \\
$\hookrightarrow$\,CrispEdit & $-0.2$ & 12.7 & 59.3 & $-1.9$ & 16.9 & 42.0 & 14.3 & 29.2 & 53.8 \\
$\hookrightarrow$\,LoRA & $-7.4$ & 22.6 & 63.1 & $-13.5$ & 18.2 & 64.6 & 31.6 & 31.1 & 53.4 \\
$\hookrightarrow$\,VILA & $-2.2$ & 19.8 & 62.9 & $-8.4$ & 17.7 & 66.8 & 29.8 & 34.8 & 59.0 \\
$\hookrightarrow$\,OPLoRA & $-4.5$ & 21.6 & 64.5 & $-15.5$ & 27.4 & 62.9 & 18.5 & 29.3 & 58.2 \\
\rowcolor{methodbg}
$\hookrightarrow$\,\mnamelora & $-1.3$ & 7.6 & 62.6 & $-3.7$ & 3.4 & 62.5 & 20.0 & 31.2 & 57.6 \\
\rowcolor{methodbg}
$\hookrightarrow$\,\mnamedense & $-0.1$ & 6.3 & 63.3 & $-1.1$ & 6.5 & 63.2 & 11.6 & 32.1 & 55.8 \\
\bottomrule
\end{tabularx}
\end{table}

In general, \mname improves the forgetting--retention balance on both TOFU and MUSE. On TOFU, \mnamedense attains the strongest Forget Quality in both forget splits and combines low forget-answer probability with model utility close to Retrain. \mnamelora further lowers forget-answer probability in the 10\% split, although its Forget Quality is lower than that of the dense variant. On MUSE, \mnamedense achieves the lowest verbatim memorization, whereas \mnamelora retains more knowledge.

The additional results show how \mname performs under GradDiff and NPO as well as SimNPO, across the three TOFU backbones and MUSE (Appendix Tables~\ref{tab:app_tofu_llama32}--\ref{tab:app_tofu_qwen3} and~\ref{tab:app_muse_full_grid}).

\subsection{Ablation Studies}
\label{sec:exp_ablations}

\begin{figure}[t]
\centering
\includegraphics[width=.94\linewidth]{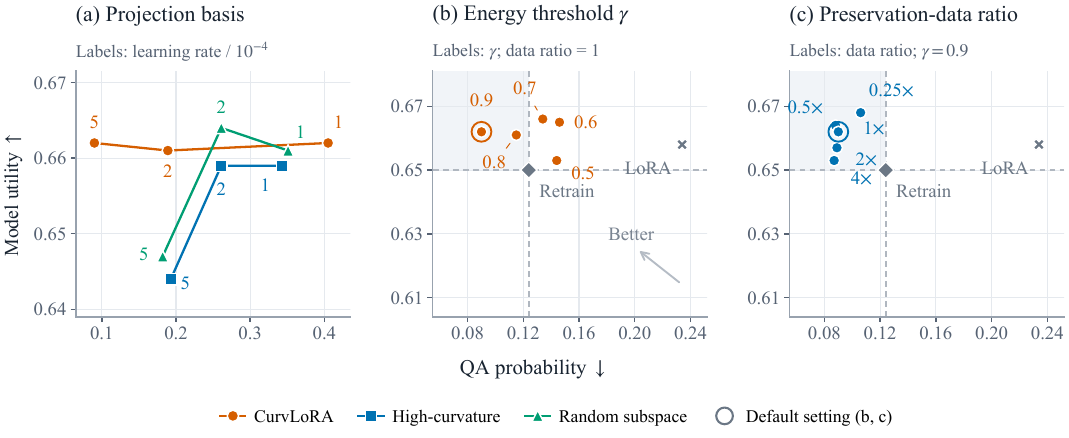}
\setlength{\abovecaptionskip}{5pt}
\caption{\textbf{Ablations of the delta-space construction.} (a) Projection-basis choices with matched dimensions; lines connect learning-rate settings, not training steps. (b) Spectral-energy threshold $\gamma$ at preservation-data ratio 1. (c) Preservation-data ratio at $\gamma=0.9$. Rings mark the default settings in (b, c). Upper left is better; (a) uses a different axis range. Dashed lines and shading mark Retrain and the region improving on it in both metrics.}
\label{fig:ablation_studies}
\end{figure}

We study four design choices on TOFU Forget 10\% (with Llama-3.2-3B-Instruct and SimNPO), chosen as a controlled setting with an exact Retrain reference. Specifically, we examine the projection basis, the spectral-energy threshold $\gamma$, and the preservation-data ratio (Figure~\ref{fig:ablation_studies}), as well as delta parameterization versus gradient projection (Figure~\ref{fig:deltagrad}).

\paragraph{Projection basis.}
To isolate the effect of direction selection, we compare our low-curvature delta space with two dimension-matched variants constructed from high-curvature and random directions. As shown in Figure~\ref{fig:ablation_studies} (a), across learning rates, updates in the low-curvature space strengthen forgetting while maintaining model utility, whereas both variants lose utility at the largest learning rate while still producing weaker forgetting. These comparisons show that constructing the delta space from preservation curvature improves the forgetting--utility trade-off beyond its dimension.

\paragraph{Spectral-energy threshold.}
The threshold $\gamma$ jointly controls the maximum retained curvature and the dimension of the available update space. As shown in Figure~\ref{fig:ablation_studies} (b), all thresholds strengthen forgetting relative to LoRA with comparable model utility; larger $\gamma$ generally further reduces QA probability while utility remains stable.

\paragraph{Preservation-data ratio.}
We estimate curvature from subsets of the retain set, with sizes from $0.25\times$ to $4\times$ that of the forget set.
As shown in Figure~\ref{fig:ablation_studies} (c), the performance of \mnamelora is robust to the amount of preservation data used for curvature estimation, indicating that our method does not require a large preservation set to estimate an effective low-curvature space.

\noindent
\begin{minipage}[t]{0.65\textwidth}
\kern0pt
\paragraph{Delta parameterization vs.\ gradient projection.}
To separate delta parameterization from subspace selection, we compare \mnamedense with a gradient-projection variant (\mbox{\mname{}Grad}) that projects task gradients with the same fixed $(\bm{P}_l^{\mathrm{out}},\bm{P}_l^{\mathrm{in}})$ before each Adam step rather than parameterizing $\Delta\bm{W}_l=\bm{P}_l^{\mathrm{out}}\bm{\delta}_l\bm{P}_l^{\mathrm{in}}$. As shown in Figure~\ref{fig:deltagrad}, \mnamedense achieves higher Forget Quality and model utility across learning rates at comparable QA probability (detailed values in Appendix Table~\ref{tab:app_deltagrad}), confirming the benefit of keeping the applied update within the low-curvature space under adaptive optimization.
\end{minipage}\hfill
\begin{minipage}[t]{0.33\textwidth}
\kern0pt
\centering
\setlength{\abovecaptionskip}{2pt}
\includegraphics[width=\linewidth]{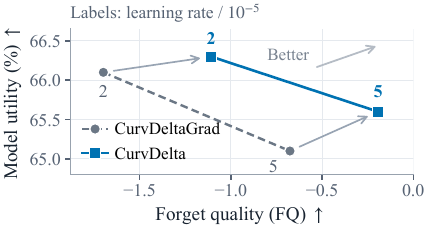}
\makeatletter\def\@captype{figure}\makeatother
\caption{\textbf{Delta parameterization vs.\ gradient projection}.}
\label{fig:deltagrad}
\end{minipage}

\section{Analysis}
\label{sec:exp_analysis}

\begin{figure}[t]
\centering
\includegraphics[width=\linewidth]{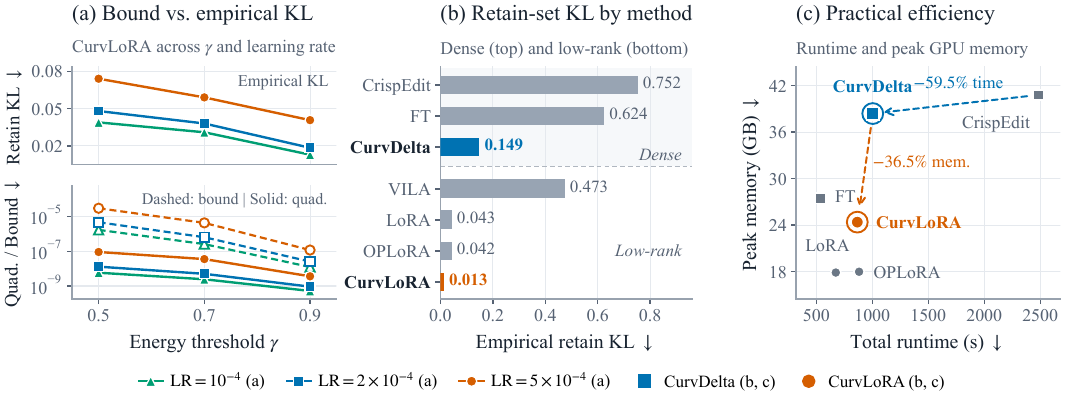}
\caption{\textbf{Preservation and efficiency analysis.} (a) Empirical retain-set KL (top) and K-FAC quadratic model with spectral bound from Eq.~\eqref{eq:method_drift_bound} (bottom) for \mnamelora. (b) Empirical retain-set KL across methods. (c) Total wall-clock runtime and peak GPU memory.}
\label{fig:analysis}
\end{figure}

\subsection{Preservation Loss and Theoretical Bound}
Continuing with the same TOFU setting, Figure~\ref{fig:analysis} (a) compares the empirical retain-set KL of \mnamelora with the K-FAC quadratic approximation and the spectral bound in Eq.~\eqref{eq:method_drift_bound} across learning rates and thresholds $\gamma$ (detailed values in Appendix Table~\ref{tab:app_kl_bound}). The K-FAC quadratic approximation remains below the spectral bound in every configuration. Although the local quadratic model around $\bm{\theta}_0$ operates on a smaller numerical scale than the full-training retain-set KL, all three quantities share the same monotonic trends: larger $\gamma$ removes more leading curvature directions and lowers all three values at each learning rate, whereas larger learning rates increase them. Across adaptation methods in Figure~\ref{fig:analysis} (b), \mnamedense yields substantially lower retain-set KL than full fine-tuning and CrispEdit, and \mnamelora attains the lowest retain-set KL among the low-rank methods.

\subsection{Computational Efficiency}
Figure~\ref{fig:analysis} (c) compares end-to-end wall-clock runtime (including curvature estimation) and peak GPU memory across methods (detailed values in Appendix Table~\ref{tab:app_efficiency}). Caching only the thin leading factor eigenbases rather than the full factor eigenbases and product mask required by CrispEdit enables \mnamedense to substantially cut total runtime with lower peak memory. Projecting low-rank factors directly in \mnamelora further reduces peak memory relative to the dense variant while matching the runtime of OPLoRA. Moreover, because the cached eigenbases depend only on the base model, preservation set, and $\gamma$, they are computed once and reused across task objectives, hyperparameters, and delta parameterizations.

\section{Conclusion}

We studied how to incorporate a preservation-loss constraint into the model delta parameterization while retaining adaptation flexibility. Our answer is \mname, which estimates local preservation curvature at the pre-adaptation model with block-diagonal K-FAC, retains trailing factor eigendirections via factorwise spectral truncation, and applies a two-sided projection so the realized delta remains in this fixed low-curvature space throughout optimization. The same geometry is reusable across tasks and objectives, supports both dense (\mnamedense) and low-rank (\mnamelora) adaptation, and yields a local preservation bound governed jointly by retained curvature and delta magnitude. Across mathematical reasoning, code generation, TOFU, and MUSE, our methods improve the acquisition--retention and forgetting--retention trade-offs over baselines, with retention close to the retraining reference.

\bibliographystyle{iclr2027_conference}
\bibliography{iclr2027_conference}


\appendix

\section{Related Works}
\label{app:related}

\paragraph{Preservation-aware task adaptation.}
Preservation-aware task adaptation addresses the challenge of acquiring capabilities required by new tasks while limiting degradation of unrelated capabilities previously acquired.
Existing approaches differ in both the signal used to identify what should be preserved and the stage of adaptation they constrain.
Replay-based methods train on stored examples from earlier tasks, while self-synthesized rehearsal generates proxy examples when the original data are unavailable \citep{rolnick2019experience,huang2024selfrehearsal}.
Optimization-based methods instead regularize parameters or reconcile competing objectives: EWC penalizes changes to parameters estimated to be important for earlier tasks \citep{kirkpatrick2017overcoming}, whereas PCGrad, CAGrad, and GradDrop can mediate conflicts between task and preservation gradients when both losses are optimized jointly \citep{yu2020gradient,liu2021conflict,chen2020just}.
PEFT methods constrain the update geometry in different ways. O-LoRA separates successive tasks through orthogonal low-rank subspaces \citep{wang2023olora}; MiLoRA adapts minor singular components of pretrained weights; and CorDA provides a knowledge-preserved decomposition intended to limit degradation of pretrained knowledge \citep{wang2025milora,yang2024corda}.
A further line of work explicitly constrains LoRA updates using subspaces derived from pretrained weights, task representations, or both \citep{xiong2026oplora,liang2024inflora,luo2026keeplora}. These constructions use orthogonality to protect selected parameter or representation subspaces during adaptation.
Model editing addresses a related preservation problem. ROME and MEMIT directly update selected MLP weights, while MEND learns an auxiliary editor that maps fine-tuning gradients to parameter updates \citep{meng2022locating,meng2023massediting,mitchell2022fast}.
AlphaEdit limits collateral changes by projecting editing perturbations into the null space of preserved-knowledge representations~\citep{fang2025alphaedit}. CrispEdit projects task gradients into a low-curvature space estimated with K-FAC~\citep{ikram2026crispedit}.
\mname derives its delta space from the local curvature of the preservation loss, relating the allowed update directions to their local preservation cost. It incorporates the projection into the delta parameterization, so that the realized delta remains in the prescribed space throughout optimization, independently of the optimizer. This combines a preservation-loss-based choice of geometry with an explicit constraint on the applied delta, supporting both dense and low-rank adaptation.

\paragraph{Selective unlearning.}
LLM unlearning seeks to suppress knowledge or behavior associated with a designated forget set while preserving utility on unrelated data, providing a practical alternative to full retraining \citep{cao2015towards}. Existing methods control this trade-off by modifying different objects in the update process. GradDiff couples ascent on the forget-set loss with training on retained data, whereas NPO and SimNPO regulate forgetting through bounded, preference-inspired losses \citep{liu2022continual,zhang2024negative,fan2025simplicity}. GRU and SAGO instead reshape the instantaneous update direction using first-order forget--retain signals: GRU projects the unlearning gradient away from directions estimated to harm retention, while SAGO treats unlearning as an asymmetric two-task problem and gates the two gradients coordinate-wise according to their sign agreement \citep{pmlr-v267-wang25de,xiao-etal-2026-modeling}. LoRA-specific methods control the content or dynamics of the low-rank update. FILA and VILA use parameter-level forget-to-retain importance to initialize adapters, whereas ReGLU uses representation geometry both to initialize a discriminative low-rank subspace and to regularize the adapter update away from retained representations \citep{cha2025fila,kim2025improving,xiao-etal-2026-representation}. Null-space-constrained low-rank adaptation also uses benign hidden representations to restrict updates while combining safe-target response learning with suppression of undesired responses \citep{ju2026nullspace}. Stable Forgetting instead applies bounded transformations to LoRA adapters to control their update dynamics \citep{garg2025stableforgetting}. SOUL is the second-order counterpart, using second-order information to shape each unlearning step \citep{jia2024soul}. \mname derives its delta space from preservation-loss curvature, providing a local loss bound in terms of the retained curvature and delta magnitude. Both \mnamedense and \mnamelora can optimize GradDiff, NPO, or SimNPO within this space.

\section{Theoretical Analysis}
\label{app:proofs}

\subsection{Bregman Divergences Satisfy Assumption~\ref{as:method_discrepancy}}
\label{app:bregman}

For a twice-differentiable convex function $\phi$ on the output space, the associated Bregman divergence is
\begin{equation}
  d_\phi(\bm{u},\bm{v})
  =\phi(\bm{u})-\phi(\bm{v})-\nabla\phi(\bm{v})^\top(\bm{u}-\bm{v}).
  \label{eq:app_bregman}
\end{equation}
Substituting $\bm{u}=\bm{v}$ gives $d_\phi(\bm{v},\bm{v})=0$. Differentiating in the first argument gives $\nabla_{\bm{u}}d_\phi(\bm{u},\bm{v})=\nabla\phi(\bm{u})-\nabla\phi(\bm{v})$, which vanishes at $\bm{u}=\bm{v}$, and $\nabla_{\bm{u}}^2d_\phi(\bm{u},\bm{v})=\nabla^2\phi(\bm{u})$, which is positive semidefinite at $\bm{u}=\bm{v}$ by convexity of $\phi$. Equation~\eqref{eq:app_bregman} therefore satisfies Assumption~\ref{as:method_discrepancy} for any such $\phi$, with no requirement that $\bm{\theta}_0$ be a stationary point of a pretraining or proxy loss.

Two instantiations are relevant here. The quadratic function $\phi(\bm{u})=\tfrac12\|\bm{u}\|_2^2$ yields $d_\phi(\bm{u},\bm{v})=\tfrac12\|\bm{u}-\bm{v}\|_2^2$. For categorical outputs with $\bm{z}$ the logit vector and $p_{\bm{z}}=\operatorname{softmax}(\bm{z})$, the log-sum-exp function $\phi(\bm{z})=\log\sum_k\exp(z_k)$ yields $d_\phi(\bm{z}_{\bm{\theta}},\bm{z}_{\bm{\theta}_0})=D_{\mathrm{KL}}(p_{\bm{\theta}_0}\|p_{\bm{\theta}})$, which is the preservation discrepancy used in all reported experiments.

\subsection{The Preservation Hessian Coincides with the GGN Matrix}
\label{app:gnh}

Write $\mathcal{L}_{\mathrm{pres}}(\bm{\theta})=\mathbb{E}_{x}[d(\bm{z}_{\bm{\theta}}(x),\bm{z}_{\bm{\theta}_0}(x))]$ and abbreviate $\bm{J}(x)=\nabla_{\bm{\theta}}\bm{z}_{\bm{\theta}}(x)$. Differentiating twice in $\bm{\theta}$ and exchanging differentiation with the expectation gives
\begin{equation}
  \nabla_{\bm{\theta}}^2\mathcal{L}_{\mathrm{pres}}(\bm{\theta})
  =\mathbb{E}_x\Bigl[
    \bm{J}(x)^\top\bigl(\nabla_{\bm{z}}^2d\bigr)\bm{J}(x)
    +\textstyle\sum_k\bigl(\partial_{z_k}d\bigr)\nabla_{\bm{\theta}}^2z_{\bm{\theta},k}(x)
  \Bigr].
  \label{eq:app_hessian_split}
\end{equation}
The first term is the generalized Gauss--Newton term. Evaluating Eq.~\eqref{eq:app_hessian_split} at $\bm{\theta}=\bm{\theta}_0$, the inner derivative $\partial_{z_k}d$ is taken at $\bm{z}=\bm{z}_{\bm{\theta}_0}(x)$, where Assumption~\ref{as:method_discrepancy} gives $\nabla_{\bm{z}}d(\bm{z},\bm{z}_{\bm{\theta}_0}(x))|_{\bm{z}=\bm{z}_{\bm{\theta}_0}(x)}=\bm{0}$. The entire sum therefore vanishes regardless of the curvature of the model outputs, leaving $\bm{H}_{\mathrm{pres}}=\mathbb{E}_x[\bm{J}_0(x)^\top \bm{H}_z(x)\bm{J}_0(x)]=\bm{G}_{\mathrm{pres}}$, which is positive semidefinite because $\bm{H}_z(x)\succeq\bm{0}$.

\subsection{Spectral Form of the Local Preservation Loss}
\label{app:preservation_spectrum}

Because $\bm{G}_{\mathrm{pres}}$ is symmetric and positive semidefinite, it admits an orthonormal eigendecomposition $\bm{G}_{\mathrm{pres}}=\bm{U}\bm{\Lambda}\bm{U}^\top$. Let $(\lambda_i,\bm{u}_i)$ denote its eigenpairs, ordered so that $\lambda_1\ge\lambda_2\ge\cdots\ge0$. Every update can be expanded in this eigenbasis as
\begin{equation*}
  \Delta\bm{\theta}
  =\sum_i\alpha_i \bm{u}_i,
  \qquad
  \alpha_i=\bm{u}_i^\top\Delta\bm{\theta}.
\end{equation*}
Using $\bm{G}_{\mathrm{pres}}\bm{u}_j=\lambda_j \bm{u}_j$ and $\bm{u}_i^\top \bm{u}_j=\mathbf{1}[i=j]$, the quadratic form becomes
\begin{align*}
  \Delta\bm{\theta}^\top \bm{G}_{\mathrm{pres}}\Delta\bm{\theta}
  &=\sum_i\sum_j
    \alpha_i\alpha_j \bm{u}_i^\top \bm{G}_{\mathrm{pres}}\bm{u}_j \\
  &=\sum_i\sum_j
    \alpha_i\alpha_j\lambda_j \bm{u}_i^\top \bm{u}_j \\
  &=\sum_i\lambda_i\alpha_i^2.
\end{align*}
Orthonormality also gives $\|\Delta\bm{\theta}\|_2^2=\sum_i\alpha_i^2$. Hence, among updates with the same norm, components along larger-eigenvalue directions contribute more to the local preservation loss. In particular,
\begin{equation*}
  \lambda_{\min}\|\Delta\bm{\theta}\|_2^2
  \le
  \Delta\bm{\theta}^\top \bm{G}_{\mathrm{pres}}\Delta\bm{\theta}
  \le
  \lambda_{\max}\|\Delta\bm{\theta}\|_2^2,
\end{equation*}
with the upper and lower bounds attained by updates aligned with leading and trailing eigenvectors, respectively.

\subsection{Kronecker Structure of the Low-Curvature Delta Space}
\label{app:low_curvature_spectrum}

This appendix derives the delta space of Section~\ref{sec:method_space} from the Kronecker eigenstructure of a K-FAC block and establishes its main properties. Consider the positive semidefinite factors with eigendecompositions $\bm{K}^{\mathrm{in}}=\bm{U}^{\mathrm{in}}\bm{\Lambda}^{\mathrm{in}}(\bm{U}^{\mathrm{in}})^\top\in\mathbb{R}^{d_{\mathrm{in}}\times d_{\mathrm{in}}}$ and $\bm{K}^{\mathrm{out}}=\bm{U}^{\mathrm{out}}\bm{\Lambda}^{\mathrm{out}}(\bm{U}^{\mathrm{out}})^\top\in\mathbb{R}^{d_{\mathrm{out}}\times d_{\mathrm{out}}}$, with eigenvalues nonnegative and ordered nonincreasingly. Let $\bm{u}_j^{\mathrm{in}}$ and $\bm{u}_i^{\mathrm{out}}$ denote the columns of $\bm{U}^{\mathrm{in}}$ and $\bm{U}^{\mathrm{out}}$. Each Kronecker eigenpair satisfies
\begin{equation}
  (\bm{K}^{\mathrm{in}}\otimes \bm{K}^{\mathrm{out}})
  (\bm{u}_j^{\mathrm{in}}\otimes \bm{u}_i^{\mathrm{out}})
  =\lambda_j^{\mathrm{in}}\lambda_i^{\mathrm{out}}
  (\bm{u}_j^{\mathrm{in}}\otimes \bm{u}_i^{\mathrm{out}}).
  \label{eq:app_kron_eigenpair}
\end{equation}
The factorwise energy cutoffs of Section~\ref{sec:method_space} retain the index set and induce the curvature coefficient
\begin{equation}
  \mathcal{R}=\{(i,j):i>r_{\mathrm{out}},\,j>r_{\mathrm{in}}\},
  \qquad
  \kappa=\lambda_{r_{\mathrm{out}}+1}^{\mathrm{out}}\lambda_{r_{\mathrm{in}}+1}^{\mathrm{in}},
  \label{eq:app_retained_directions}
\end{equation}
with the boundary convention $\lambda_{d_{\mathrm{out}}+1}^{\mathrm{out}}=\lambda_{d_{\mathrm{in}}+1}^{\mathrm{in}}=0$.

For any retained subset of index pairs with binary indicator mask $\bm{M}\in\{0,1\}^{d_{\mathrm{out}}\times d_{\mathrm{in}}}$, expanding $\operatorname{vec}(\bm{\delta})$ in the orthonormal basis of Eq.~\eqref{eq:app_kron_eigenpair} gives the orthogonal projection of $\bm{\delta}\in\mathbb{R}^{d_{\mathrm{out}}\times d_{\mathrm{in}}}$ onto the corresponding Kronecker eigenspace~\citep{ikram2026crispedit}:
\begin{equation}
  \bm{\Pi}_{\bm{M}}(\bm{\delta})
  =\bm{U}^{\mathrm{out}}
  \Bigl(
    \bigl((\bm{U}^{\mathrm{out}})^\top\bm{\delta}\bm{U}^{\mathrm{in}}\bigr)
    \odot\bm{M}
  \Bigr)
  (\bm{U}^{\mathrm{in}})^\top.
  \label{eq:app_exact_product_projection}
\end{equation}
For the factorwise index set $\mathcal{R}$, the indicator entries factor as $M_{ij}=\mathbf{1}[i>r_{\mathrm{out}}]\mathbf{1}[j>r_{\mathrm{in}}]$. Define the binary indicator vectors $\bm{m}^{\mathrm{out}}\in\{0,1\}^{d_{\mathrm{out}}}$ and $\bm{m}^{\mathrm{in}}\in\{0,1\}^{d_{\mathrm{in}}}$ by $m_i^{\mathrm{out}}=\mathbf{1}[i>r_{\mathrm{out}}]$ and $m_j^{\mathrm{in}}=\mathbf{1}[j>r_{\mathrm{in}}]$, so that $\bm{M}=\bm{m}^{\mathrm{out}}(\bm{m}^{\mathrm{in}})^\top$. Because Hadamard multiplication of any matrix $\bm{C}$ by $\bm{m}^{\mathrm{out}}(\bm{m}^{\mathrm{in}})^\top$ equals $\operatorname{diag}(\bm{m}^{\mathrm{out}})\,\bm{C}\,\operatorname{diag}(\bm{m}^{\mathrm{in}})$, substituting $\bm{M}=\bm{m}^{\mathrm{out}}(\bm{m}^{\mathrm{in}})^\top$ into Eq.~\eqref{eq:app_exact_product_projection} and regrouping into left- and right-multiplications on $\bm{\delta}$ yields the separable form
\begin{align}
  \bm{\Pi}_{\bm{M}}(\bm{\delta})
  &=\bm{U}^{\mathrm{out}}
  \Bigl(
    \bigl((\bm{U}^{\mathrm{out}})^\top\bm{\delta}\bm{U}^{\mathrm{in}}\bigr)
    \odot\bigl(\bm{m}^{\mathrm{out}}(\bm{m}^{\mathrm{in}})^\top\bigr)
  \Bigr)
  (\bm{U}^{\mathrm{in}})^\top \nonumber\\
  &=\bm{U}^{\mathrm{out}}
  \Bigl(
    \operatorname{diag}(\bm{m}^{\mathrm{out}})
    (\bm{U}^{\mathrm{out}})^\top\bm{\delta}\bm{U}^{\mathrm{in}}
    \operatorname{diag}(\bm{m}^{\mathrm{in}})
  \Bigr)
  (\bm{U}^{\mathrm{in}})^\top \nonumber\\
  &=\bigl(\bm{U}^{\mathrm{out}}\operatorname{diag}(\bm{m}^{\mathrm{out}})(\bm{U}^{\mathrm{out}})^\top\bigr)
  \,\bm{\delta}\,
  \bigl(\bm{U}^{\mathrm{in}}\operatorname{diag}(\bm{m}^{\mathrm{in}})(\bm{U}^{\mathrm{in}})^\top\bigr).
  \label{eq:app_separable_projection}
\end{align}
Partition the orthonormal eigenbases into leading and trailing blocks as $\bm{U}^{\mathrm{out}}=[\bm{U}_h^{\mathrm{out}},\bm{U}_t^{\mathrm{out}}]$ and $\bm{U}^{\mathrm{in}}=[\bm{U}_h^{\mathrm{in}},\bm{U}_t^{\mathrm{in}}]$, where $\bm{U}_h^{\mathrm{out}}\in\mathbb{R}^{d_{\mathrm{out}}\times r_{\mathrm{out}}}$ and $\bm{U}_h^{\mathrm{in}}\in\mathbb{R}^{d_{\mathrm{in}}\times r_{\mathrm{in}}}$ collect the leading eigenvectors. The diagonal $0$--$1$ masks select the trailing blocks, and completeness of the orthonormal bases ($\bm{U}_h^{\mathrm{out}}(\bm{U}_h^{\mathrm{out}})^\top+\bm{U}_t^{\mathrm{out}}(\bm{U}_t^{\mathrm{out}})^\top=\bm{I}$ and $\bm{U}_h^{\mathrm{in}}(\bm{U}_h^{\mathrm{in}})^\top+\bm{U}_t^{\mathrm{in}}(\bm{U}_t^{\mathrm{in}})^\top=\bm{I}$) rewrites the left and right projectors via complementary subspace projection:
\begin{equation}
  \begin{aligned}
    \bm{U}^{\mathrm{out}}\operatorname{diag}(\bm{m}^{\mathrm{out}})(\bm{U}^{\mathrm{out}})^\top
    &=\bm{U}_t^{\mathrm{out}}(\bm{U}_t^{\mathrm{out}})^\top
    =\bm{I}-\bm{U}_h^{\mathrm{out}}(\bm{U}_h^{\mathrm{out}})^\top
    =\bm{P}^{\mathrm{out}}, \\
    \bm{U}^{\mathrm{in}}\operatorname{diag}(\bm{m}^{\mathrm{in}})(\bm{U}^{\mathrm{in}})^\top
    &=\bm{U}_t^{\mathrm{in}}(\bm{U}_t^{\mathrm{in}})^\top
    =\bm{I}-\bm{U}_h^{\mathrm{in}}(\bm{U}_h^{\mathrm{in}})^\top
    =\bm{P}^{\mathrm{in}}.
  \end{aligned}
  \label{eq:app_complementary_projectors}
\end{equation}
Substituting Eq.~\eqref{eq:app_complementary_projectors} into Eq.~\eqref{eq:app_separable_projection} recovers the projectors in Eq.~\eqref{eq:method_delta_projectors} and the two-sided projection $\Delta\bm{W}=\bm{P}^{\mathrm{out}}\bm{\delta}\bm{P}^{\mathrm{in}}$ in Eq.~\eqref{eq:method_delta_projection} (or $\operatorname{vec}(\Delta\bm{W})=(\bm{P}^{\mathrm{in}}\otimes\bm{P}^{\mathrm{out}})\operatorname{vec}(\bm{\delta})$ in vectorized form). Thus, $\bm{\delta}\mapsto\bm{P}^{\mathrm{out}}\bm{\delta}\bm{P}^{\mathrm{in}}$ is the orthogonal projector onto the span of $\{\bm{u}_j^{\mathrm{in}}\otimes \bm{u}_i^{\mathrm{out}}:(i,j)\in\mathcal{R}\}$. The retained space has dimension $|\mathcal{R}|=(d_{\mathrm{out}}-r_{\mathrm{out}})(d_{\mathrm{in}}-r_{\mathrm{in}})$, its largest eigenvalue is $\kappa$ whenever nonempty, and the projection preserves rank bounds: $\operatorname{rank}(\bm{P}^{\mathrm{out}}\bm{\delta}\bm{P}^{\mathrm{in}})\le\operatorname{rank}(\bm{\delta})$. Because $r_{\mathrm{out}}(\gamma)$ and $r_{\mathrm{in}}(\gamma)$ are nondecreasing in $\gamma$ and the factor eigenvalues are ordered nonincreasingly, both $\kappa$ and $|\mathcal{R}|$ are nonincreasing in $\gamma$.

\subsection{Relation to the Full Product-Thresholded Eigenspace}
\label{app:product_threshold}

Rather than truncating each factor spectrum separately, CrispEdit~\citep{ikram2026crispedit} thresholds the K-FAC product spectrum directly. For a curvature threshold $\tau\ge0$, the product-thresholded index set is
\begin{equation}
  \mathcal{R}^{\mathrm{prod}}_\tau
  =\{(i,j):\lambda_i^{\mathrm{out}}\lambda_j^{\mathrm{in}}\le\tau\}.
  \label{eq:app_product_threshold_set}
\end{equation}
With indicator mask $\bm{M}_\tau\in\{0,1\}^{d_{\mathrm{out}}\times d_{\mathrm{in}}}$ defined by $(M_\tau)_{ij}=\mathbf{1}[(i,j)\in\mathcal{R}^{\mathrm{prod}}_\tau]$, the orthogonal projection onto this eigenspace is $\bm{\Pi}_{\bm{M}_\tau}(\bm{\delta})$ in Eq.~\eqref{eq:app_exact_product_projection}, whose range contains every Kronecker eigendirection with eigenvalue at most $\tau$.

Setting $\tau=\kappa$ relates the two index sets: every $(i,j)\in\mathcal{R}$ satisfies $\lambda_i^{\mathrm{out}}\lambda_j^{\mathrm{in}}\le\kappa$, so
\begin{equation}
  \mathcal{R}\subseteq\mathcal{R}^{\mathrm{prod}}_\kappa.
  \label{eq:app_separable_subset}
\end{equation}
The inclusion can be strict because $\mathcal{R}^{\mathrm{prod}}_\kappa$ also admits mixed pairs with $i\le r_{\mathrm{out}}$ or $j\le r_{\mathrm{in}}$ whenever a sufficiently small eigenvalue in one factor offsets a large eigenvalue in the other.

However, a generic $\bm{M}_\tau$ is not a rank-$1$ outer product of an output mask and an input mask, so $\bm{\Pi}_{\bm{M}_\tau}(\bm{\delta})$ in Eq.~\eqref{eq:app_exact_product_projection} cannot be separated into left and right multiplications as in Eq.~\eqref{eq:app_separable_projection}. Applying $\bm{\Pi}_{\bm{M}_\tau}$ therefore requires caching the full factor eigenbases $\bm{U}^{\mathrm{out}}$ and $\bm{U}^{\mathrm{in}}$ along with $\bm{M}_\tau$, transforming into and out of the full bases at each step, and applying an elementwise mask that can increase the rank of $\bm{\delta}$ (Table~\ref{tab:app_complexity}).

\subsection{Low-Curvature Delta Control}
\label{app:low_curvature_control}

The Kronecker structure established in Appendix~\ref{app:low_curvature_spectrum} bounds the quadratic cost of each projected layer delta. The following result depends only on the two positive semidefinite factors and their retained eigenspaces.

\begin{proposition}[Low-curvature delta control]
\label{app:prop_low_curvature}
Let $\mathcal{R}$ and $\kappa$ be the retained direction set and curvature coefficient of Eq.~\eqref{eq:app_retained_directions}. Every applied delta $\Delta\bm{W}=\bm{P}^{\mathrm{out}}\bm{\delta}\bm{P}^{\mathrm{in}}$ satisfies
\begin{equation}
  \operatorname{vec}(\Delta\bm{W})^\top
  (\bm{K}^{\mathrm{in}}\otimes \bm{K}^{\mathrm{out}})
  \operatorname{vec}(\Delta\bm{W})
  \le
  \kappa\|\Delta\bm{W}\|_F^2.
  \label{eq:app_low_curvature_bound}
\end{equation}
\end{proposition}

\begin{proof}
By Appendix~\ref{app:low_curvature_spectrum}, the map $\bm{\delta}\mapsto \bm{P}^{\mathrm{out}}\bm{\delta}\bm{P}^{\mathrm{in}}$ is the orthogonal projector onto the span of the Kronecker eigenvectors $\bm{u}_j^{\mathrm{in}}\otimes \bm{u}_i^{\mathrm{out}}$ with $(i,j)\in\mathcal{R}$. Expanding $\operatorname{vec}(\Delta\bm{W})=\sum_{(i,j)\in\mathcal{R}}c_{ij}\,(\bm{u}_j^{\mathrm{in}}\otimes \bm{u}_i^{\mathrm{out}})$ in this orthonormal basis,
\begin{equation*}
  \operatorname{vec}(\Delta\bm{W})^\top(\bm{K}^{\mathrm{in}}\otimes \bm{K}^{\mathrm{out}})\operatorname{vec}(\Delta\bm{W})
  =\sum_{(i,j)\in\mathcal{R}}\lambda_i^{\mathrm{out}}\lambda_j^{\mathrm{in}}\,c_{ij}^2
  \le\kappa\sum_{(i,j)\in\mathcal{R}}c_{ij}^2
  =\kappa\|\Delta\bm{W}\|_F^2,
\end{equation*}
where the inequality uses that the eigenvalues are sorted in descending order, so $\lambda_i^{\mathrm{out}}\le\lambda_{r_{\mathrm{out}}+1}^{\mathrm{out}}$ for $i>r_{\mathrm{out}}$ and $\lambda_j^{\mathrm{in}}\le\lambda_{r_{\mathrm{in}}+1}^{\mathrm{in}}$ for $j>r_{\mathrm{in}}$.
\end{proof}

When the retained span is nonempty, the bound is attained by a direction with eigenvalue $\kappa$.

\subsection{Proof of Theorem~\ref{thm:method_epsilon_feasible}}
\label{app:thm_epsilon}

To aggregate the layerwise bounds into the full preservation loss, we use the following model of the curvature on the adapted weights.

\begin{assumption}[Block-diagonal curvature]
\label{app:as_blockdiag}
The preservation GGN quadratic form on updates supported on $\mathcal{T}$ follows the block-diagonal K-FAC model, with blocks $\bm{K}_l^{\mathrm{in}}\otimes\bm{K}_l^{\mathrm{out}}$ for $l\in\mathcal{T}$. All weights outside $\mathcal{T}$ remain unchanged.
\end{assumption}

\begin{proof}
Assumption~\ref{as:method_discrepancy} gives the local quadratic approximation in Eq.~\eqref{eq:method_preservation_quadratic}, and Appendix~\ref{app:gnh} identifies its Hessian with $\bm{G}_{\mathrm{pres}}$. Under Assumption~\ref{app:as_blockdiag},
\begin{equation*}
  \begin{aligned}
    \Delta\bm{\theta}^\top\bm{G}_{\mathrm{pres}}\Delta\bm{\theta}
    &=\sum_{l\in\mathcal{T}}
    \operatorname{vec}(\Delta\bm{W}_l)^\top
    (\bm{K}_l^{\mathrm{in}}\otimes\bm{K}_l^{\mathrm{out}})
    \operatorname{vec}(\Delta\bm{W}_l)\\
    &\le\sum_{l\in\mathcal{T}}\kappa_l\|\Delta\bm{W}_l\|_F^2,
  \end{aligned}
\end{equation*}
where the inequality applies Proposition~\ref{app:prop_low_curvature} to each block. Multiplying by $1/2$ gives Eq.~\eqref{eq:method_drift_bound}. Comparing this bound with $\varepsilon$ establishes the sufficient condition under the local quadratic model.
\end{proof}

\section{Curvature Estimation and Computational Cost}
\label{app:kfac_details}

\subsection{Monte Carlo Fisher Estimation}
\label{app:mc_fisher}

We instantiate the preservation discrepancy with token-level $D_{\mathrm{KL}}(p_{\bm{\theta}_0}\|p_{\bm{\theta}})$. Up to a term independent of $\bm{\theta}$, this is the expected negative log-likelihood under pseudo-targets drawn from the original model. Following standard Monte Carlo Fisher estimation in K-FAC \citep{martens2015optimizing}, for each preservation context we sample $\widetilde y_t\sim p_{\bm{\theta}_0}(\cdot\mid x_{<t})$ and backpropagate
\begin{equation}
  \ell_{\mathrm{MC}}(\bm{\theta};x)
  =-\sum_t\log p_{\bm{\theta}}(\widetilde y_t\mid x_{<t}).
  \label{eq:app_mc_fisher_loss}
\end{equation}
At $\bm{\theta}_0$, the expected outer product of the scores of Eq.~\eqref{eq:app_mc_fisher_loss} is the model Fisher, which for softmax outputs with a KL discrepancy equals the GGN matrix in Eq.~\eqref{eq:method_preservation_gnh}. Accumulating the input activations and pre-activation pseudo-gradients of this backward pass yields the factors $\bm{K}_l^{\mathrm{in}}$ and $\bm{K}_l^{\mathrm{out}}$ of Eq.~\eqref{eq:method_kfac}.

\subsection{Complexity Accounting}
\label{app:complexity}

\mname constructs the preservation geometry once and applies the resulting projector during task optimization. Both \mnamedense and \mnamelora cache the same thin eigenvector bases; their projection costs differ with the delta parameterization. We account for construction and application separately, with all layerwise costs stated for $\bm{W}_l\in\mathbb{R}^{d_{\mathrm{out}}\times d_{\mathrm{in}}}$.

\paragraph{One-time construction and caching.}
Estimating the K-FAC factors requires one forward and backward pass over $\mathcal{D}_{\mathrm{pres}}$ under Eq.~\eqref{eq:app_mc_fisher_loss}. Per target weight, the factors occupy $O(d_{\mathrm{out}}^2+d_{\mathrm{in}}^2)$ memory and their eigendecompositions cost $O(d_{\mathrm{out}}^3+d_{\mathrm{in}}^3)$. After selecting the energy cutoffs, we discard the dense factors and full eigenbases, retaining only $\bm{U}_h^{\mathrm{out}}$, $\bm{U}_h^{\mathrm{in}}$, and $\kappa_l$. With the base model, preservation set, target weights, and threshold $\gamma$ fixed, this cache is reusable across tasks, optimization hyperparameters, and both delta parameterizations.

For comparison, the exact preservation GGN for an $N$-parameter model requires $O(N^2)$ storage and $O(N^3)$ eigendecomposition time; even one exact weight block costs $O(d_{\mathrm{out}}^3d_{\mathrm{in}}^3)$ to diagonalize. The product-thresholded projector of Eq.~\eqref{eq:app_exact_product_projection} uses the same K-FAC eigendecompositions as \mname but retains the full factor eigenbases and a product mask. Table~\ref{tab:app_complexity} compares the resulting caches and application costs, and Figure~\ref{fig:analysis} (c) reports the corresponding empirical runtime and peak memory measurements.

\begin{table}[ht]
\centering
\small
\renewcommand{\arraystretch}{1.15}
\caption{Per-weight storage and cost of one projection. Cached geometry excludes trainable deltas and temporary buffers. Application is to a dense delta except for \mnamelora, which projects rank-$r$ factors.}
\label{tab:app_complexity}
\begin{tabularx}{\textwidth}{@{}L{3.2cm}YY@{}}
\toprule
\textbf{Projection} & \textbf{Cache storage} & \textbf{Projection cost} \\
\midrule
Exact GGN eigenspace & $O(d_{\mathrm{out}}^2d_{\mathrm{in}}^2)$ & $O(d_{\mathrm{out}}^2d_{\mathrm{in}}^2)$ \\
Product threshold \citep{ikram2026crispedit} & $O(d_{\mathrm{out}}^2+d_{\mathrm{in}}^2+d_{\mathrm{out}}d_{\mathrm{in}})$ & $\Theta(d_{\mathrm{out}}d_{\mathrm{in}}(d_{\mathrm{out}}+d_{\mathrm{in}}))$ \\
\midrule
\rowcolor{methodbg}
\mnamedense (dense) & $O(d_{\mathrm{out}}r_{\mathrm{out}}+d_{\mathrm{in}}r_{\mathrm{in}})$ & $\Theta(d_{\mathrm{out}}d_{\mathrm{in}}(r_{\mathrm{out}}+r_{\mathrm{in}}))$ \\
\rowcolor{methodbg}
\mnamelora (rank $r$) & $O(d_{\mathrm{out}}r_{\mathrm{out}}+d_{\mathrm{in}}r_{\mathrm{in}})$ & $O(r(d_{\mathrm{out}}r_{\mathrm{out}}+d_{\mathrm{in}}r_{\mathrm{in}}))$ \\
\bottomrule
\end{tabularx}
\end{table}

\paragraph{Projection during optimization.}
For \mnamedense, we apply Eq.~\eqref{eq:method_delta_projection} through four multiplications by the cached thin bases:
\begin{equation*}
\begin{aligned}
\bm{D}_l
&=\bm{\delta}_l-(\bm{\delta}_l\bm{U}_h^{\mathrm{in}})
  (\bm{U}_h^{\mathrm{in}})^\top,\\
\Delta\bm{W}_l
&=\bm{D}_l-\bm{U}_h^{\mathrm{out}}
  ((\bm{U}_h^{\mathrm{out}})^\top\bm{D}_l).
\end{aligned}
\end{equation*}
This gives the dense projection cost in Table~\ref{tab:app_complexity}. For \mnamelora, the factorization in Section~\ref{sec:method_algorithm} instead applies the same thin-basis operations to $\widetilde{\bm{B}}_l=\bm{P}^{\mathrm{out}}\bm{B}_l$ and $\widetilde{\bm{A}}_l=\bm{A}_l\bm{P}^{\mathrm{in}}$, reducing projection work when $r\ll\min(d_{\mathrm{out}},d_{\mathrm{in}})$. The projected factors occupy $O(r(d_{\mathrm{out}}+d_{\mathrm{in}}))$ memory, and evaluating $\widetilde{\bm{B}}_l(\widetilde{\bm{A}}_l\bm{a}_{l-1})$ costs $O(r(d_{\mathrm{out}}+d_{\mathrm{in}}))$ per input vector without materializing a dense delta. The table reports projection work only, rather than the cost of a full training step.

\section{Unlearning Objectives}
\label{app:objectives}

In this section, we instantiate the general constrained update problem in Eq.~\eqref{eq:prelim_constrained} for the unlearning experiments. Here $\mathcal{D}_{\mathrm{task}}=\mathcal{D}_f$ is the forget set and $\mathcal{D}_{\mathrm{pres}}=\mathcal{D}_r$ is the retain set. Let $\ell(\bm{\theta};x)=-\sum_{t=1}^{T}\log p_{\bm{\theta}}(x_t\mid x_{<t})$ denote the negative log-likelihood of a sequence $x$.

\paragraph{Gradient Difference (GradDiff).} GradDiff \citep{liu2022continual,maini2024tofu} combines gradient ascent on the forget-set NLL with gradient descent on the retain-set NLL. Written as a minimization objective, it is
\begin{equation}
  \mathcal{L}_{\mathrm{GradDiff}}(\bm{\theta})
  =
  -\mathbb{E}_{x\sim\mathcal{D}_f}\bigl[\ell(\bm{\theta};x)\bigr]
  +
  \alpha\,
  \mathbb{E}_{x\sim\mathcal{D}_r}\bigl[\ell(\bm{\theta};x)\bigr],
  \label{eq:prelim_graddiff}
\end{equation}
where $\alpha>0$ controls the contribution of retain-set supervision.

\paragraph{Negative Preference Optimization (NPO).} NPO \citep{zhang2024negative} replaces unbounded ascent with a reference-anchored preference objective:
\begin{equation}
  \mathcal{L}_{\mathrm{NPO}}(\bm{\theta};\mathcal{D}_f)
  =
  -\frac{2}{\beta}\,
  \mathbb{E}_{x\sim\mathcal{D}_f}
  \log\sigma\!\left(
    -\beta\log\frac{p_{\bm{\theta}}(x)}{p_{\bm{\theta}_0}(x)}
  \right),
  \label{eq:prelim_npo}
\end{equation}
where $\sigma$ is the logistic sigmoid and $\beta>0$ controls the sharpness of the bounded transformation.

\paragraph{SimNPO.} SimNPO \citep{fan2025simplicity} removes the reference-model ratio from NPO and applies the bounded transformation to the length-normalized forget-sequence log-likelihood:
\begin{equation}
  \mathcal{L}_{\mathrm{SimNPO}}(\bm{\theta};\mathcal{D}_f)
  =
  -\frac{2}{\beta}\,
  \mathbb{E}_{x\sim\mathcal{D}_f}
  \log\sigma\!\left(
    -\frac{\beta}{|x|}\log p_{\bm{\theta}}(x)-\gamma_{\mathrm{SimNPO}}
  \right),
  \label{eq:prelim_simnpo}
\end{equation}
where $|x|$ is the sequence length and $\gamma_{\mathrm{SimNPO}}$ is the margin controlling the target level of forget-sequence suppression.

\section{Experimental Details}
\label{app:experimental_details}

\subsection{Capability Learning}
\label{app:learning_training_details}

Section~\ref{sec:exp_learning_setup} summarizes the datasets and evaluation setup; here, we provide complementary details about preservation set and hyperparameter selection. For the preservation set, we sample 256 preservation examples from the FineFineWeb validation split \citep{map2024finefineweb}. For training, we retrain LoRA, PiSSA, and MiLoRA using the configurations selected by \citet{lee2026learningrate}. For FT, Replay, CorDA, OPLoRA, \mnamedense, and \mnamelora, we tune the learning rate over the grid in Table~\ref{tab:app_capability_training}; the remaining settings are fixed across methods.
\begin{table}[H]
\centering
\small
\caption{Training hyperparameters and search space for capability learning.}
\label{tab:app_capability_training}
\begin{tabularx}{\textwidth}{@{}C{4.0cm}Y@{}}
\toprule
\textbf{Hyperparameter} & \textbf{Value / search space} \\
\midrule
Training epochs & 1 \\
Optimizer & AdamW \\
Learning-rate schedule & Cosine decay \\
Warmup ratio & 0.03 \\
Weight decay & 0 \\
Dropout & 0 \\
Maximum sequence length & 512 \\
Effective batch size & 64 (Gemma-3-1B); 16 (Llama-2-7B) \\
LoRA rank & 128 (Gemma-3-1B); 32 (Llama-2-7B) \\
LoRA target modules & \texttt{\{q,k,v,o,gate,up,down\}\_proj} \\
Learning rate &
$\{1.12,2.00,3.56,6.32\}\times\{10^{-5},10^{-4}\}$ \\
\bottomrule
\end{tabularx}
\end{table}

\subsection{Selective Unlearning}
\label{app:unlearning_training_details}

Section~\ref{sec:exp_unlearning_setup} summarizes the benchmarks, models, objectives, and evaluation setup; here, we provide complementary details about data splits, curvature construction, and hyperparameter selection.

\subsubsection{TOFU}

We use the OpenUnlearning implementation.\footnote{\url{https://github.com/locuslab/open-unlearning}} Table~\ref{tab:app_tofu_training} lists the fixed settings and search spaces used for experiments.

\begin{table}[H]
\centering
\small
\caption{Training hyperparameters and search space on TOFU. Settings are shared across the three backbones and both forget splits.}
\label{tab:app_tofu_training}
\begin{tabularx}{\textwidth}{@{}C{4.0cm}Y@{}}
\toprule
\textbf{Hyperparameter} & \textbf{Value / search space} \\
\midrule
Training epochs & 10 \\
Optimizer & \texttt{adamw\_8bit} \\
Effective batch size & 32 \\
Warmup epochs & 1 \\
Weight decay & 0.01 \\
LoRA rank & 32 \\
LoRA scaling factor & 64 \\
LoRA dropout & 0.05 \\
LoRA target modules & \texttt{\{q,k,v,o,gate,up,down\}\_proj} \\
Curvature-energy ratio ($\gamma$) & $\{0.5,0.6,0.7,0.8,0.9\}$ \\
\bottomrule
\end{tabularx}
\end{table}

\subsubsection{MUSE}

Table~\ref{tab:app_muse_training} lists the training settings shared across objectives. We search the learning rate over $\{1,2,5\}\times10^{-5}$ for GradDiff, which has no temperature parameter. For NPO, we search learning rates in $\{2,5,10\}\times10^{-5}$ and temperatures $\beta\in\{0.005,0.01\}$. For SimNPO, we search learning rates in $\{0.5,1,2\}\times10^{-4}$ and the same temperature range $\beta\in\{0.005,0.01\}$.

\begin{table}[H]
\centering
\small
\caption{Shared training settings for FT, LoRA, and \mnamelora on MUSE Books.}
\label{tab:app_muse_training}
\begin{tabularx}{\textwidth}{@{}C{4.0cm}Y@{}}
\toprule
\textbf{Shared hyperparameter} & \textbf{Value} \\
\midrule
Training epochs & 5 \\
Optimizer & \texttt{adamw\_8bit} \\
Effective batch size & 32 \\
LoRA rank & 32 \\
LoRA scaling factor & 64 \\
\bottomrule
\end{tabularx}
\end{table}

\section{Additional Results}
\label{app:additional_results}

This appendix reports the complete results across unlearning objectives and methods underlying the selective-unlearning comparisons in Section~\ref{sec:exp_unlearning_results}, as well as the numerical values underlying the parameterization ablation in Section~\ref{sec:exp_ablations} and the preservation and efficiency analysis in Section~\ref{sec:exp_analysis}.

\subsection{Complete TOFU Results}
\label{app:main_tofu_grid}

Tables~\ref{tab:app_tofu_llama32}--\ref{tab:app_tofu_qwen3} report the complete TOFU grid for Llama-3.2-3B-Instruct, Llama-3.1-8B-Instruct, and Qwen3-4B-Instruct-2507. Each table covers two forget splits and three losses (GradDiff, NPO, and SimNPO). As in Table~\ref{tab:main_unlearning}, Retrain and Original are reference rows, then FT, CrispEdit, LoRA, VILA, OPLoRA, \mnamelora, and \mnamedense are nested under each objective. Training configurations are specified in Appendix~\ref{app:unlearning_training_details}.

\begin{table}[H]
\centering
\scriptsize
\setlength{\tabcolsep}{3.2pt}
\caption{Complete TOFU results on Llama-3.2-3B-Instruct. Retrain and Original are reference rows; subsequent rows are grouped by unlearning objective.}
\label{tab:app_tofu_llama32}
\begin{tabularx}{\textwidth}{l *{6}{Y}}
\toprule
\multirow{2}{*}{\textbf{Method}} & \multicolumn{3}{c}{\textbf{Forget 5\%}} & \multicolumn{3}{c}{\textbf{Forget 10\%}} \\
\cmidrule(lr){2-4}\cmidrule(lr){5-7}
& \mbox{FQ $\uparrow$} & \mbox{FP $\downarrow$} & \mbox{MU $\uparrow$} & \mbox{FQ $\uparrow$} & \mbox{FP $\downarrow$} & \mbox{MU $\uparrow$} \\
\midrule
\rowcolor{basebg}
Retrain & -- & 13.1 & 65.9 & -- & 12.4 & 65.0 \\
\rowcolor{basebg}
Original & $-13.6$ & 95.1 & 66.6 & $-26.1$ & 95.1 & 66.6 \\
\cmidrule(lr){1-7}
\multicolumn{7}{l}{GradDiff} \\
$\hookrightarrow$\,FT & $-102.5$ & 0.0 & 56.9 & $-220.4$ & 0.0 & 39.2 \\
$\hookrightarrow$\,CrispEdit & $-100.9$ & 0.0 & 54.8 & $-202.6$ & 0.0 & 48.1 \\
$\hookrightarrow$\,LoRA & $-12.5$ & 85.1 & 66.1 & $-26.4$ & 11.6 & 51.3 \\
$\hookrightarrow$\,VILA & $-11.8$ & 12.4 & 56.2 & $-26.1$ & 6.7 & 54.0 \\
$\hookrightarrow$\,OPLoRA & $-99.3$ & 0.0 & 60.2 & $-220.4$ & 0.0 & 64.1 \\
\rowcolor{methodbg}
$\hookrightarrow$\,\mnamelora & $-100.8$ & 0.0 & 67.7 & $-173.7$ & 3.1 & 67.1 \\
\rowcolor{methodbg}
$\hookrightarrow$\,\mnamedense & $-8.2$ & 4.7 & 66.8 & $-173.8$ & 0.7 & 67.1 \\
\cmidrule(lr){1-7}
\multicolumn{7}{l}{NPO} \\
$\hookrightarrow$\,FT & $-1.9$ & 10.3 & 47.4 & $-2.1$ & 4.7 & 50.4 \\
$\hookrightarrow$\,CrispEdit & $-1.9$ & 4.0 & 60.4 & $-1.1$ & 3.6 & 57.0 \\
$\hookrightarrow$\,LoRA & $-14.7$ & 91.2 & 66.7 & $-18.4$ & 58.2 & 59.0 \\
$\hookrightarrow$\,VILA & $-2.0$ & 11.2 & 38.3 & $-9.0$ & 15.5 & 41.6 \\
$\hookrightarrow$\,OPLoRA & $-0.8$ & 17.2 & 50.0 & $-0.7$ & 7.9 & 64.5 \\
\rowcolor{methodbg}
$\hookrightarrow$\,\mnamelora & $-0.4$ & 13.4 & 66.2 & $-0.2$ & 2.4 & 67.4 \\
\rowcolor{methodbg}
$\hookrightarrow$\,\mnamedense & $-1.2$ & 4.0 & 66.3 & $-0.2$ & 9.1 & 66.5 \\
\cmidrule(lr){1-7}
\multicolumn{7}{l}{SimNPO} \\
$\hookrightarrow$\,FT & $-0.1$ & 16.2 & 56.6 & $-1.2$ & 16.8 & 58.1 \\
$\hookrightarrow$\,CrispEdit & $-0.5$ & 17.3 & 60.2 & $-0.9$ & 19.2 & 61.5 \\
$\hookrightarrow$\,LoRA & $-10.2$ & 53.5 & 65.4 & $-10.9$ & 23.4 & 65.8 \\
$\hookrightarrow$\,VILA & $-1.8$ & 25.3 & 55.7 & $-7.0$ & 23.2 & 65.6 \\
$\hookrightarrow$\,OPLoRA & $-6.1$ & 28.0 & 65.3 & $-11.3$ & 22.9 & 65.6 \\
\rowcolor{methodbg}
$\hookrightarrow$\,\mnamelora & $-1.2$ & 3.7 & 66.4 & $-2.3$ & 13.4 & 66.6 \\
\rowcolor{methodbg}
$\hookrightarrow$\,\mnamedense & $-0.1$ & 10.9 & 66.1 & $-0.6$ & 9.1 & 65.7 \\
\bottomrule
\end{tabularx}
\end{table}

\begin{table}[H]
\centering
\scriptsize
\setlength{\tabcolsep}{3.2pt}
\caption{Complete TOFU results on Llama-3.1-8B-Instruct. Retrain and Original are reference rows; subsequent rows are grouped by unlearning objective.}
\label{tab:app_tofu_llama31}
\begin{tabularx}{\textwidth}{l *{6}{Y}}
\toprule
\multirow{2}{*}{\textbf{Method}} & \multicolumn{3}{c}{\textbf{Forget 5\%}} & \multicolumn{3}{c}{\textbf{Forget 10\%}} \\
\cmidrule(lr){2-4}\cmidrule(lr){5-7}
& \mbox{FQ $\uparrow$} & \mbox{FP $\downarrow$} & \mbox{MU $\uparrow$} & \mbox{FQ $\uparrow$} & \mbox{FP $\downarrow$} & \mbox{MU $\uparrow$} \\
\midrule
\rowcolor{basebg}
Retrain & -- & 10.8 & 63.2 & -- & 10.6 & 64.6 \\
\rowcolor{basebg}
Original & $-11.8$ & 99.0 & 62.7 & $-26.8$ & 99.2 & 62.7 \\
\cmidrule(lr){1-7}
\multicolumn{7}{l}{GradDiff} \\
$\hookrightarrow$\,FT & $-109.7$ & 0.0 & 43.8 & $-23.0$ & 0.0 & 0.0 \\
$\hookrightarrow$\,CrispEdit & $-111.7$ & 0.0 & 59.8 & $-190.3$ & 1.2 & 47.5 \\
$\hookrightarrow$\,LoRA & $-14.0$ & 9.3 & 48.1 & $-38.7$ & 17.0 & 47.3 \\
$\hookrightarrow$\,VILA & $-11.2$ & 9.9 & 51.4 & $-37.4$ & 11.3 & 52.4 \\
$\hookrightarrow$\,OPLoRA & $-100.9$ & 0.0 & 58.2 & $-204.2$ & 0.0 & 60.6 \\
\rowcolor{methodbg}
$\hookrightarrow$\,\mnamelora & $-118.7$ & 0.0 & 62.0 & $-236.0$ & 0.0 & 63.3 \\
\rowcolor{methodbg}
$\hookrightarrow$\,\mnamedense & $-100.9$ & 0.0 & 63.1 & $-228.7$ & 0.0 & 65.5 \\
\cmidrule(lr){1-7}
\multicolumn{7}{l}{NPO} \\
$\hookrightarrow$\,FT & $-1.7$ & 7.0 & 54.1 & $-1.0$ & 7.3 & 57.2 \\
$\hookrightarrow$\,CrispEdit & $-4.1$ & 2.0 & 63.3 & $-1.0$ & 3.5 & 60.7 \\
$\hookrightarrow$\,LoRA & $-12.9$ & 97.2 & 62.7 & $-26.1$ & 72.8 & 58.5 \\
$\hookrightarrow$\,VILA & $-4.5$ & 51.5 & 59.5 & $-9.8$ & 18.1 & 26.7 \\
$\hookrightarrow$\,OPLoRA & $-0.6$ & 14.5 & 61.9 & $-8.6$ & 5.4 & 68.0 \\
\rowcolor{methodbg}
$\hookrightarrow$\,\mnamelora & $-0.9$ & 2.3 & 61.4 & $-2.3$ & 2.9 & 66.9 \\
\rowcolor{methodbg}
$\hookrightarrow$\,\mnamedense & $-0.3$ & 5.6 & 62.5 & $-1.6$ & 4.7 & 63.7 \\
\cmidrule(lr){1-7}
\multicolumn{7}{l}{SimNPO} \\
$\hookrightarrow$\,FT & $-0.5$ & 13.0 & 54.6 & $-1.4$ & 14.4 & 38.9 \\
$\hookrightarrow$\,CrispEdit & $-0.2$ & 12.7 & 59.3 & $-1.9$ & 16.9 & 42.0 \\
$\hookrightarrow$\,LoRA & $-7.4$ & 22.6 & 63.1 & $-13.5$ & 18.2 & 64.6 \\
$\hookrightarrow$\,VILA & $-2.2$ & 19.8 & 62.9 & $-8.4$ & 17.7 & 66.8 \\
$\hookrightarrow$\,OPLoRA & $-4.5$ & 21.6 & 64.5 & $-15.5$ & 27.4 & 62.9 \\
\rowcolor{methodbg}
$\hookrightarrow$\,\mnamelora & $-1.3$ & 7.6 & 62.6 & $-3.7$ & 3.4 & 62.5 \\
\rowcolor{methodbg}
$\hookrightarrow$\,\mnamedense & $-0.1$ & 6.3 & 63.3 & $-1.1$ & 6.5 & 63.2 \\
\bottomrule
\end{tabularx}
\end{table}

\begin{table}[H]
\centering
\scriptsize
\setlength{\tabcolsep}{3.2pt}
\caption{Complete TOFU results on Qwen3-4B-Instruct-2507. Retrain and Original are reference rows; subsequent rows are grouped by unlearning objective.}
\label{tab:app_tofu_qwen3}
\begin{tabularx}{\textwidth}{l *{6}{Y}}
\toprule
\multirow{2}{*}{\textbf{Method}} & \multicolumn{3}{c}{\textbf{Forget 5\%}} & \multicolumn{3}{c}{\textbf{Forget 10\%}} \\
\cmidrule(lr){2-4}\cmidrule(lr){5-7}
& \mbox{FQ $\uparrow$} & \mbox{FP $\downarrow$} & \mbox{MU $\uparrow$} & \mbox{FQ $\uparrow$} & \mbox{FP $\downarrow$} & \mbox{MU $\uparrow$} \\
\midrule
\rowcolor{basebg}
Retrain & -- & 11.5 & 58.5 & -- & 11.1 & 59.7 \\
\rowcolor{basebg}
Original & $-10.9$ & 98.8 & 56.0 & $-20.8$ & 98.9 & 56.0 \\
\cmidrule(lr){1-7}
\multicolumn{7}{l}{GradDiff} \\
$\hookrightarrow$\,FT & $-3.0$ & 0.1 & 51.0 & $-220.4$ & 0.0 & 39.2 \\
$\hookrightarrow$\,CrispEdit & $-104.2$ & 0.0 & 60.5 & $-233.5$ & 0.0 & 58.9 \\
$\hookrightarrow$\,LoRA & $-6.3$ & 55.8 & 56.0 & $-22.7$ & 1.5 & 43.9 \\
$\hookrightarrow$\,VILA & $-8.2$ & 24.5 & 54.7 & $-20.5$ & 26.6 & 55.6 \\
$\hookrightarrow$\,OPLoRA & $-11.2$ & 6.1 & 46.8 & $-220.3$ & 0.0 & 56.9 \\
\rowcolor{methodbg}
$\hookrightarrow$\,\mnamelora & $-82.7$ & 4.9 & 54.6 & $-147.6$ & 6.4 & 56.6 \\
\rowcolor{methodbg}
$\hookrightarrow$\,\mnamedense & $-2.5$ & 1.4 & 56.7 & $-30.1$ & 8.1 & 58.0 \\
\cmidrule(lr){1-7}
\multicolumn{7}{l}{NPO} \\
$\hookrightarrow$\,FT & $-2.2$ & 5.1 & 45.5 & $-0.1$ & 5.2 & 47.2 \\
$\hookrightarrow$\,CrispEdit & $-0.5$ & 3.6 & 51.9 & $-0.6$ & 2.7 & 52.7 \\
$\hookrightarrow$\,LoRA & $-5.9$ & 21.3 & 53.0 & $-9.8$ & 10.8 & 56.3 \\
$\hookrightarrow$\,VILA & $-10.2$ & 91.2 & 56.4 & $-13.7$ & 0.0 & 0.3 \\
$\hookrightarrow$\,OPLoRA & $-3.4$ & 23.9 & 50.0 & $-5.5$ & 11.1 & 56.5 \\
\rowcolor{methodbg}
$\hookrightarrow$\,\mnamelora & $-2.2$ & 12.0 & 57.7 & $-1.3$ & 4.7 & 59.6 \\
\rowcolor{methodbg}
$\hookrightarrow$\,\mnamedense & $-0.1$ & 3.3 & 55.3 & $-0.1$ & 3.9 & 55.7 \\
\cmidrule(lr){1-7}
\multicolumn{7}{l}{SimNPO} \\
$\hookrightarrow$\,FT & $-1.2$ & 18.2 & 44.7 & $-2.2$ & 16.5 & 51.0 \\
$\hookrightarrow$\,CrispEdit & $-1.1$ & 14.6 & 52.3 & $-7.0$ & 16.1 & 54.9 \\
$\hookrightarrow$\,LoRA & $-8.2$ & 48.1 & 57.7 & $-16.4$ & 38.9 & 58.2 \\
$\hookrightarrow$\,VILA & $-8.2$ & 51.5 & 57.9 & $-15.3$ & 32.7 & 57.2 \\
$\hookrightarrow$\,OPLoRA & $-7.9$ & 36.6 & 58.4 & $-15.3$ & 29.4 & 59.4 \\
\rowcolor{methodbg}
$\hookrightarrow$\,\mnamelora & $-1.4$ & 10.6 & 57.0 & $-4.7$ & 15.4 & 57.0 \\
\rowcolor{methodbg}
$\hookrightarrow$\,\mnamedense & $-0.3$ & 8.1 & 55.5 & $-0.0$ & 8.8 & 54.9 \\
\bottomrule
\end{tabularx}
\end{table}

\subsection{Complete MUSE Results}
\label{app:main_muse_grid}

Table~\ref{tab:app_muse_full_grid} reports every combination used to support the MUSE Books comparison in the main paper. As in Table~\ref{tab:main_unlearning}, Retrain and Original are reference rows, then FT, CrispEdit, LoRA, VILA, OPLoRA, \mnamelora, and \mnamedense are nested under each objective.

\begin{table}[H]
\centering
\scriptsize
\setlength{\tabcolsep}{3.2pt}
\caption{Complete MUSE Books results with Llama-2-7B. Retrain and Original are reference rows; subsequent rows are grouped by unlearning objective.}
\label{tab:app_muse_full_grid}
\begin{tabularx}{0.85\textwidth}{l *{3}{Y}}
\toprule
\multirow{2}{*}{\textbf{Method}} & \multicolumn{2}{c}{\textbf{Forget}} & \mbox{\textbf{Retain}} \\
\cmidrule(lr){2-3}\cmidrule(lr){4-4}
& \mbox{VerbMem $\downarrow$} & \mbox{KnowMem $\downarrow$} & \mbox{KnowMem $\uparrow$} \\
\midrule
\rowcolor{basebg}
Retrain & 14.2 & 29.5 & 68.6 \\
\rowcolor{basebg}
Original & 99.7 & 46.6 & 69.1 \\
\cmidrule(lr){1-4}
\multicolumn{4}{l}{GradDiff} \\
$\hookrightarrow$\,FT & 0.0 & 0.0 & 0.0 \\
$\hookrightarrow$\,CrispEdit & 88.7 & 39.6 & 60.8 \\
$\hookrightarrow$\,LoRA & 89.9 & 40.3 & 57.9 \\
$\hookrightarrow$\,VILA & 0.6 & 5.1 & 21.6 \\
$\hookrightarrow$\,OPLoRA & 88.5 & 37.8 & 57.3 \\
\rowcolor{methodbg}
$\hookrightarrow$\,\mnamelora & 20.9 & 28.3 & 61.1 \\
\rowcolor{methodbg}
$\hookrightarrow$\,\mnamedense & 0.5 & 39.7 & 65.5 \\
\cmidrule(lr){1-4}
\multicolumn{4}{l}{NPO} \\
$\hookrightarrow$\,FT & 12.3 & 2.9 & 2.2 \\
$\hookrightarrow$\,CrispEdit & 13.3 & 29.0 & 59.2 \\
$\hookrightarrow$\,LoRA & 27.7 & 38.1 & 60.9 \\
$\hookrightarrow$\,VILA & 97.3 & 40.8 & 64.0 \\
$\hookrightarrow$\,OPLoRA & 26.5 & 29.9 & 53.2 \\

\rowcolor{methodbg}
$\hookrightarrow$\,\mnamelora & 26.5 & 29.5 & 60.5 \\
\rowcolor{methodbg}
$\hookrightarrow$\,\mnamedense & 6.3 & 16.2 & 51.3 \\
\cmidrule(lr){1-4}
\multicolumn{4}{l}{SimNPO} \\
$\hookrightarrow$\,FT & 23.3 & 27.6 & 46.8 \\
$\hookrightarrow$\,CrispEdit & 14.3 & 29.2 & 53.8 \\
$\hookrightarrow$\,LoRA & 31.6 & 31.1 & 53.4 \\
$\hookrightarrow$\,VILA & 29.8 & 34.8 & 59.0 \\
$\hookrightarrow$\,OPLoRA & 18.5 & 29.3 & 58.2 \\
\rowcolor{methodbg}
$\hookrightarrow$\,\mnamelora & 20.0 & 31.2 & 57.6 \\
\rowcolor{methodbg}
$\hookrightarrow$\,\mnamedense & 11.6 & 32.1 & 55.8 \\
\bottomrule
\end{tabularx}
\end{table}

\subsection{Detailed Values for Parameterization Ablation}
\label{app:deltagrad_details}

Table~\ref{tab:app_deltagrad} reports the numerical values plotted in Figure~\ref{fig:deltagrad} comparing \mnamedense with gradient projection (\mname{}Grad) under the same low-curvature projectors on TOFU Forget 10\% with Llama-3.2-3B-Instruct and SimNPO.

\begin{table}[H]
\centering
\small
\setlength{\tabcolsep}{4pt}
\caption{Comparison of \mnamedense and \mname{}Grad under the same low-curvature projectors on TOFU Forget 10\% with Llama-3.2-3B-Instruct and SimNPO (underlying Figure~\ref{fig:deltagrad}).}
\label{tab:app_deltagrad}
\begin{tabularx}{\textwidth}{l l Y Y Y}
\toprule
\textbf{Learning Rate} & \textbf{Method} & \mbox{\textbf{FQ} $\uparrow$} & \mbox{\textbf{FP (\%)} $\downarrow$} & \mbox{\textbf{MU (\%)} $\uparrow$} \\
\midrule
$2\times 10^{-5}$ & \mname{}Grad & $-1.699$ & 11.8 & 66.1 \\
\rowcolor{methodbg}
$2\times 10^{-5}$ & \mnamedense & \textbf{$-1.108$} & \textbf{11.3} & \textbf{66.3} \\
\midrule
$5\times 10^{-5}$ & \mname{}Grad & $-0.676$ & \textbf{10.0} & 65.1 \\
\rowcolor{methodbg}
$5\times 10^{-5}$ & \mnamedense & \textbf{$-0.193$} & 10.2 & \textbf{65.6} \\
\bottomrule
\end{tabularx}
\end{table}

\clearpage
\subsection{Detailed Values for Preservation and Efficiency Analysis}
\label{app:kl_bound_details}

Tables~\ref{tab:app_kl_bound} and~\ref{tab:app_efficiency} report the numerical measurements plotted in Figure~\ref{fig:analysis} on TOFU Forget 10\% with Llama-3.2-3B-Instruct and SimNPO (runtime and peak GPU memory measured on an NVIDIA L40 GPU with a per-device batch size of 4).

\begin{table}[H]
\centering
\small
\setlength{\tabcolsep}{4pt}
\caption{Empirical retain-set KL, K-FAC quadratic approximation, and spectral bound from Eq.~\eqref{eq:method_drift_bound} for \mnamelora across learning rates and thresholds $\gamma$ on TOFU Forget 10\% with Llama-3.2-3B-Instruct and SimNPO (underlying Figure~\ref{fig:analysis}a).}
\label{tab:app_kl_bound}
\begin{tabularx}{\textwidth}{c c Y Y Y}
\toprule
\textbf{Learning Rate} & $\bm{\gamma}$ & \mbox{\textbf{Empirical KL} $\downarrow$} & \mbox{\textbf{K-FAC Quadratic} $\downarrow$} & \mbox{\textbf{Spectral Bound} $\downarrow$} \\
\midrule
$10^{-4}$ & 0.5 & 0.03892 & $5.82\times 10^{-9}$ & $1.72\times 10^{-6}$ \\
$10^{-4}$ & 0.7 & 0.03097 & $2.44\times 10^{-9}$ & $2.60\times 10^{-7}$ \\
$10^{-4}$ & 0.9 & 0.01296 & $5.17\times 10^{-10}$ & $1.24\times 10^{-8}$ \\
\midrule
$2\times 10^{-4}$ & 0.5 & 0.04803 & $1.30\times 10^{-8}$ & $4.73\times 10^{-6}$ \\
$2\times 10^{-4}$ & 0.7 & 0.03803 & $5.03\times 10^{-9}$ & $6.48\times 10^{-7}$ \\
$2\times 10^{-4}$ & 0.9 & 0.01853 & $9.14\times 10^{-10}$ & $2.56\times 10^{-8}$ \\
\midrule
$5\times 10^{-4}$ & 0.5 & 0.07398 & $9.12\times 10^{-8}$ & $3.04\times 10^{-5}$ \\
$5\times 10^{-4}$ & 0.7 & 0.05899 & $3.56\times 10^{-8}$ & $4.36\times 10^{-6}$ \\
$5\times 10^{-4}$ & 0.9 & 0.04072 & $3.62\times 10^{-9}$ & $1.18\times 10^{-7}$ \\
\bottomrule
\end{tabularx}
\end{table}

\begin{table}[H]
\centering
\small
\setlength{\tabcolsep}{4pt}
\caption{Retain-set KL, total wall-clock runtime (including curvature estimation), and peak GPU memory across adaptation methods on TOFU Forget 10\% with Llama-3.2-3B-Instruct and SimNPO (underlying Figure~\ref{fig:analysis}b, c).}
\label{tab:app_efficiency}
\begin{tabularx}{\textwidth}{l l Y Y Y}
\toprule
\textbf{Type} & \textbf{Method} & \mbox{\textbf{Retain KL} $\downarrow$} & \mbox{\textbf{Runtime (s)} $\downarrow$} & \mbox{\textbf{Peak Memory (GB)} $\downarrow$} \\
\midrule
Dense & FT & 0.62394 & 537 & 27.5 \\
Dense & CrispEdit & 0.75187 & 2482 & 40.8 \\
\rowcolor{methodbg}
Dense & \mnamedense & \textbf{0.14866} & 1004 & 38.4 \\
\midrule
Low-rank & LoRA & 0.04302 & 674 & 17.9 \\
Low-rank & VILA & 0.47349 & -- & -- \\
Low-rank & OPLoRA & 0.04222 & 882 & 18.0 \\
\rowcolor{methodbg}
Low-rank & \mnamelora & \textbf{0.01296} & 866 & 24.4 \\
\bottomrule
\end{tabularx}
\end{table}



\end{document}